\documentclass[lettersize,journal]{IEEEtran}
\usepackage{amsmath,amsfonts}
\usepackage{algorithmic}
\usepackage{algorithm}
\usepackage{array}
\usepackage[caption=false,font=normalsize,labelfont=sf,textfont=sf]{subfig}
\usepackage{textcomp}
\usepackage{stfloats}
\usepackage{url}
\usepackage{verbatim}
\usepackage{graphicx}
\usepackage{cite}
\usepackage{tabularx}
\usepackage{multirow}
\usepackage{xcolor}
\usepackage{booktabs}
\usepackage{tabularx}
\usepackage{makecell}

\begin{document}

\title{FIIH: Fully Invertible Image Hiding for Secure and Robust}

\author{Lang Huang, Lin Huo, Zheng Gan, Xinrong He,
\thanks{Lang Huang and Lin Huo contributed equally to this research.}
\thanks{Corresponding Author: Lin Huo}
\thanks{Lang Huang are with the School of Computer and Electronic Information,Guangxi University,Nanning 530004,China(e-mail:806062224@qq.com;)}
\thanks{Lin Huo are with the China-ASEAN Research Institute,Guangxi University,Nanning 530004,China(e-mail:Lhuo@gxu.edu.cn;)}
\thanks{Xinrong He, Mathematics Group of Qinzhou No.1 Middle School,Qinzhou 535099,China(e-mail:hexinronghxr@163.com)}
\thanks{Zheng Gan are with the Guangxi Zhuang Autonomous Region Information Center,Nanning 530201,China}
\thanks{This work was supported by Open Project Program of Guangxi Key Laboratory of Digital Infrastructure (Project Number:GXDIOP2023007).}
}

\markboth{Journal of \LaTeX\ Class Files,~Vol.~14, No.~8, August~2021}%
{}


\maketitle

\begin{abstract}
Image hiding is the study of techniques for covert storage and transmission, which embeds a secret image into a container image and generates stego image to make it similar in appearance to a normal image. However, existing image hiding methods have a serious problem that the hiding and revealing process cannot be fully invertible, which results in the revealing network not being able to recover the secret image losslessly, which makes it impossible to simultaneously achieve high fidelity and secure transmission of the secret image in an insecure network environment. To solve this problem,this paper proposes a fully invertible image hiding architecture based on invertible neural network,aiming to realize invertible hiding of secret images,which is invertible on both data and network. Based on this ingenious architecture, the method can withstand deep learning based image steganalysis. In addition, we propose a new method for enhancing the robustness of stego images after interference during transmission. Experiments demonstrate that the FIIH proposed in this paper significantly outperforms other state-of-the-art image hiding methods in hiding a single image, and also significantly outperforms other state-of-the-art methods in robustness and security.
\end{abstract}


\section{Introduction}
\IEEEPARstart{W}{ith} the rapid development of digital communication and artificial intelligence, digital watermarking \cite{hsu1999hidden}, copyright protection and privacy protection \cite{abraham2004significance} are increasingly in the spotlight, and these applications involve image hiding techniques. Image hiding, as a widely researched technique \cite{johnson1998exploring,morkel2005overview,cheddad2010digital}, aims to ensure that secret information can be received undetected by embedding it in digital images, while resisting noise interference and image steganography analysis during transmission. Unlike cryptography, steganography does not require secret communication, but rather the transmission of the ciphertext in the open. On the contrary, image hiding cleverly hides the information in ordinary images, and does not want anyone to notice any difference between the stego image and the ordinary image in order to realize the secret communication, which makes image hiding more covert.

Traditional image hiding methods can be mainly categorized into three main groups: null-domain hiding \cite{chan2004hiding,tsai2009reversible,wu2003steganographic,pan2011image,das2012novel,imaizumi2014multibit}, transform-domain hiding \cite{fridrich2007statistically,hetzl2005graph,provos2003hide,sallee2003model,fridrich2001detecting,hayes2017generating,hsu1999hidden} and adaptive image hiding \cite{holub2014universal,holub2012designing,pevny2010using}. Null domain hiding is achieved by embedding the secret message into the pixel values of the image, e.g., Chan \cite{chan2004hiding} employs LSB steganography to hide the secret message in the lowest bit.Das \cite{das2012novel}, on the other hand, employs Hoffman coding to convert the secret message into a binary code, which is subsequently embedded in the cover image using the LSB method. Transform domain hiding, on the other hand, embeds the secret message into the transform domain of the image, such as frequency domain and wavelet domain, to improve the robustness of the image. In transform domain hiding, secret information is hidden in the frequency domain by modifying the frequency coefficients of the transformed image, such as Discrete Fourier Transform (DFT) \cite{fridrich2001detecting}, Discrete Cosine Transform (DCT) \cite{hayes2017generating} and Discrete Wavelet Transform (DWT) \cite{hsu1999hidden}. These methods effectively improve the image hiding robustness by transforming the information from the null domain to the frequency domain. Adaptive image hiding, on the other hand, focuses on adaptively selecting the region of hidden information by analyzing the texture complexity.Tomas \cite{pevny2010using} minimizes the distortion by employing an effective coding algorithm that defines the distortion as the weighted difference of the latest feature vectors extended in the steganalysis. However, although these traditional image hiding methods have achieved some results in information hiding, they have limited hiding capacity. These methods mainly rely on hand-crafted features that are easily detected by steganalysis algorithms, resulting in relatively low security and small capacity of the resulting stego images.

\begin{figure}[!t]
\centering
\includegraphics[width=3.5in]{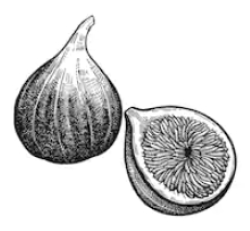}
\caption{illustrates the difference between our image hiding method and traditional invertible neural network methods.}
\label{fig_1}
\end{figure}

Convolutional Neural Networks (CNNs) have achieved remarkable results in several fields, and deep learning based image hiding methods have come to the forefront as a result.Baluja \cite{baluja2017hiding} proposed a convolutional neural network based image hiding method, which includes three networks, namely, preparation network, hiding network, and reveal network.Wu et al \cite{wu2018image} and \cite{wu2018stegnet} designed an encoder-decoder architecture based on U-Net for hiding.Goodfellow et al \cite{goodfellow2014generative} proposed a Generative Adversarial Network (GAN) based on game theory to realize the image generation task through the game of generator and discriminator networks. In this GAN framework, the two networks compete with each other to generate high-quality images.Volkhonskiy et al \cite{volkhonskiy2016generative} proposed the SGAN model in 2017, which uses a random noise input to generate a realistic carrier image through DCGAN \cite{radford2015unsupervised}, then hides the secret information through a traditional ±1 embedding algorithm, and finally inputs the secret-containing image into a steganalysis network, which is trained through adversarial training to improve the anti-detection ability. In 2020, Zhang \cite{zhang2020udh} proposed the method of summing the secret image and cover image for image hiding and isolated the encoding of the secret image from the cover image, proving that the success of deep learning based image hiding can be attributed to the frequency difference between the cover image and the encoded secret image.Baluja \cite{baluja2019hiding} proposed a system consisting of three deep neural networks that achieve embedding of multiple full color images in a single image. Although methods based on convolutional neural networks have been able to hide multiple images in a single image, the hiding and revealing networks used in the above methods are not invertible, while a large amount of information is lost in the extracted secret images. To solve these problems, invertible neural network has been introduced.

In recent years, image hiding methods based on invertible neural networks \cite{jing2021hinet,lu2021large,guan2022deepmih,shang2023robust} have achieved remarkable results. As shown in Fig. 1(a), in the past methods, the hiding network and the revealing network share the same parameters, realizing the invertibility of the hiding network and the extracting network. However, it is difficult to directly utilize the loss information r in the revealing process because the loss information r outputted by the hiding process is agnostic at the receiver side.These methods use a constant matrix or a random Gaussian as an auxiliary variable, but the loss information r is not the same as that of the constant matrix or the random Gaussian in terms of content and distribution. In \cite{xu2022robust,huo2024fitting}, although the loss information r is utilized, the generated auxiliary variable z is never equal to r, which leads to the degradation of the quality of the extracted secret image. To solve this problem, the image hiding method proposed in this paper is shown in Fig. 1(b), which not only realizes the reversibility of the hiding network and the extraction network, but also realizes the invertible hiding of the data. To the best of my knowledge, previous methods have failed to achieve invertible hiding of network and data. Moreover, the method is resistant to image steganalysis based on deep learning. In addition, this paper proposes methods to enhance the image for its robustness. As a result, The FIIH proposed in this paper achieves state-of-the-art performance in terms of security, robustness resistance and invisibility of stego images, and state-of-the-art performance in terms of fidelity of secret images. The important contributions of this paper are summarized as follows:

We find that image hiding methods based on invertible neural networks fail to achieve fully invertible hiding of data, and lay the foundation for realizing fully invertible networks and data.

We propose a new architecture based on invertible neural networks that successfully realizes fully invertible image hiding techniques. This innovation drives the quality of stego images and extracted secret images to a new level.

Through our innovation of proposing a method that does not add the secret image to the cover image, we are able to improve the resistance to image steganalysis, thus ensuring security during transmission.

We propose a method to improve image robustness. Through the enhancement module, we effectively deal with the interference during the transmission of the stego image and aim to improve the robustness of the stego image.

The rest of the paper is organized as follows. Section 2 presents a series of existing papers on robust image hiding and image hiding based on invertible neural networks. Section 3 describes our proposed method in detail. Subsequently, in Section 4 we evaluate the performance of the proposed method. Finally, Section 5 summarizes the paper and gives an outlook.

\section{Related Works}
\subsection{Robust Image Hiding}
Robustness refers to the ability of stego images to maintain good performance in the face of attacks such as image processing, compression algorithms, and noise. In related research, Yu \cite{yu2020robust} proposed a robust steganography method incorporating jitter modulation, which constructs asymmetric costs with different correction polarities to enhance the robustness of image hiding.Zhang \cite{zhang2021image} proposed an image robust adaptive steganography method for secret communication in open social networks, which possesses robustness against image processing and resistance to detection.Zhao \cite{zhao2018improving} proposed a method that utilizes transmission channel matching to resist JPEG compression and improve the robustness of adaptive steganography.Zhu \cite{zhu2018hidden} proposed the HiDDeN framework, which is a learnable end-to-end image steganography and watermarking model. It consists of an encoder, a parameter-free noise layer, a decoder, and an adversarial discriminator, and is designed to improve the robustness of JPEG images.Yu \cite{yu2020attention} proposed an Attention-Based Mechanism for Data Hiding (ABDH) method, which introduces an attentional module in the end-to-end framework. During the training process, inconsistent loss and periodic discriminative models are introduced to improve the image quality of the generated target images. In addition, Shang \cite{shang2023robust} proposed an end-to-end robust data hiding scheme for JPEG images that utilizes a bidirectional process invertible neural network to embed and extract secret messages on the quantized DCT coefficients, which is inherently robust to lossy JPEG compression. A JPEG compression attack module is designed to simulate the JPEG compression process. With the training of the JPEG compression attack module, the network is able to automatically learn how to recover the embedded secret messages from the attacked target image. Although these data hiding methods achieve robustness to a certain extent, they suffer from the problems of limited hiding capacity, not being able to hide the image sufficiently, and low robustness.

Xu \cite{xu2022robust} proposed a novel stream-based framework called RIIS, which employs conditionally normalized streams to model the distribution of redundant high-frequency components in the container image condition. The method also introduces a container enhancement module (CEM) to improve robust reconstruction. By using distortion-guided modulation (DGM), the network parameters are tuned to accommodate different distortion levels. The goal is to prevent distortions such as Poisson noise, Gaussian noise and JPEG compression from container images during hiding. On the other hand, Yu \cite{yu2023cross} proposed the CRoSS architecture, which introduces the diffusion model into the image hiding domain for the first time. The framework takes full advantage of the unique properties of the diffusion model and aims to improve the performance such as security, controllability and robustness. Although the above methods are able to resist certain noise attacks while hiding images, their effectiveness still needs to be improved when facing attacks such as Gaussian noise, JPEG compression and dropout. In order to cope with these challenges, this paper proposes an innovative approach against dropout attacks and successfully resists Gaussian noise and JPEG compression by the innovative architecture proposed in this paper. This innovation is expected to further improve the robustness of image hiding methods in complex attack scenarios.

\begin{figure*}[!t]
\centering
\includegraphics[width=7in]{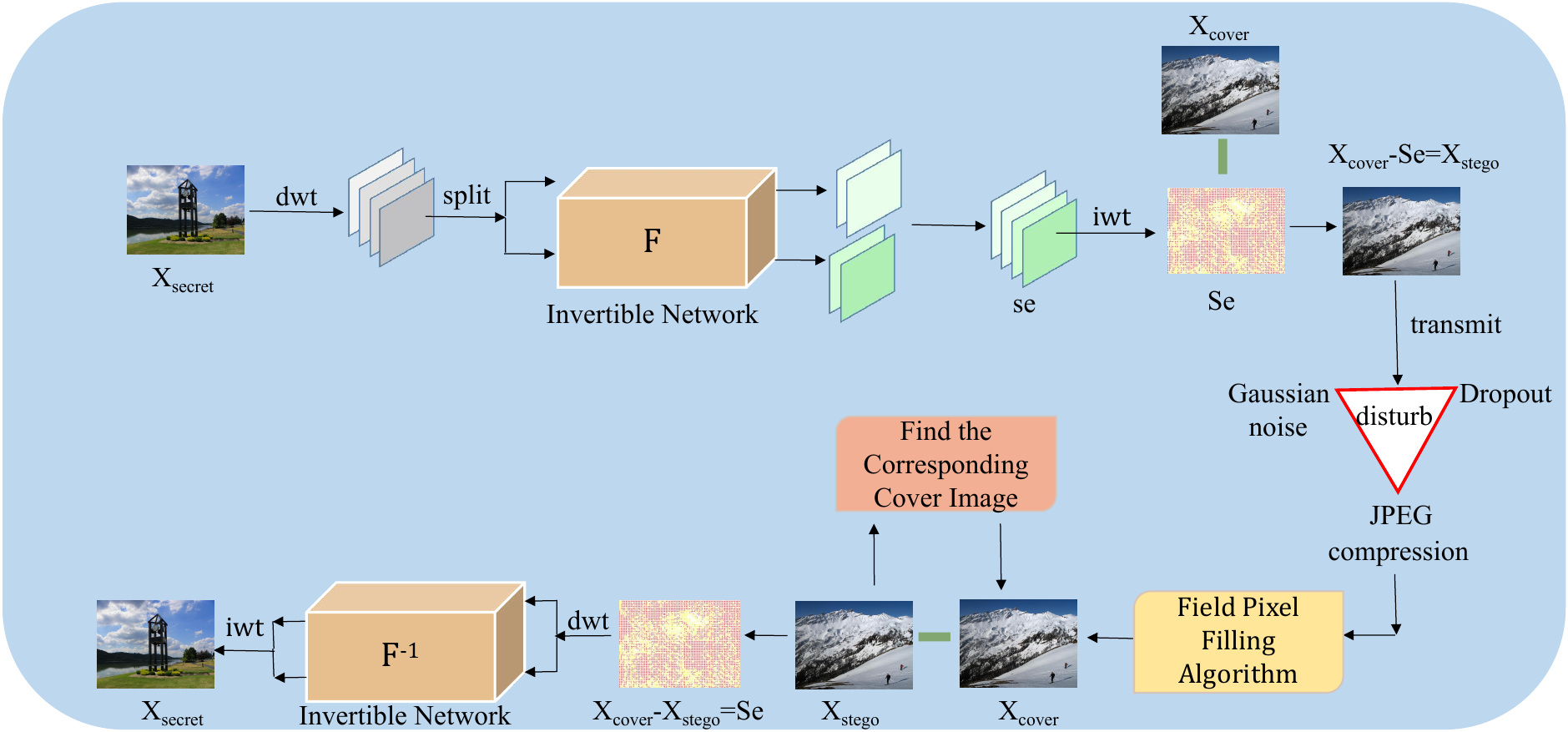} 
\caption{Architecture of FIIH's overall model.}
\label{fig_2}
\end{figure*}

\subsection{Invertible Neural Network}
Invertible neural networks (INNs) have made rapid development in various fields in recent years and are widely used in specific applications such as computer vision and image processing. Its unique feature is that it allows forward and backward computation through neural networks, which provides a strong support for tasks in several fields.Dinh et al \cite{dinh2014nice} proposed invertible neural networks (INN) for the first time and introduced convolutional and multiscale layers in the coupled model in order to reduce the computational cost.Kingma et al \cite{kingma2018glow} proposed the Glow model, and in order to enhance the performance of the invertible neural network, the Reversible 1×1 convolution was introduced.Ouderaa et al \cite{van2019reversible} explored approximate invertible architectures by using pix2pix and cycleGan loss functions to solve image-to-image conversion tasks.Xiao et al \cite{xiao2020invertible} attempted to utilize invertible neural networks for image reconstruction to recover high-frequency information and details from scaled-down low-resolution images.Lugmayr et al \cite{lugmayr2020srflow} used normalized flow model for generating high resolution images.Wang et al \cite{wang2020modeling} applied invertible neural network for image compression task. In addition, Xu et al \cite{liu2021invertible} proposed InvDN for image denoising using invertible neural network.

Invertible neural networks have been widely used in image hiding tasks 
\cite{jing2021hinet,lu2021large,guan2022deepmih,shang2023robust,xu2022robust,huo2024fitting}. These studies have proposed different methods for image hiding. In \cite{jing2021hinet}, Jing et al. for the first time applied invertible neural networks to the image hiding task of a single image and proposed a novel framework called HiNet, which hides secret information in the wavelet transform domain. To improve the security of hiding, a new low-frequency wavelet loss is introduced to constrain the secret information in the high-frequency wavelet sub-band. In \cite{lu2021large}, Lu et al. implemented the hiding of multiple graphs using invertible neural network to achieve large capacity steganography. In \cite{guan2022deepmih}, Guan et al. used an invertible neural network framework called DeepMiH for multi-image hiding, and designed an importance mapping (IM) module to utilize the previous image hiding results to guide the current image hiding, which effectively improved the performance of multi-image hiding. However, the above method fails to handle the loss information r well in the hiding process, resulting in low quality of the generated images.

In Xu's study \cite{xu2022robust}, a novel stream-based framework called RIIS is proposed, aiming to solve the imperfection problem when using Gaussian distribution or constant matrix as inputs.RIIS proposes an innovative solution: at the sender side, the loss information r is inputted into an invertible neural network to generate a Gaussian distribution with the dependency of the stego image; at the receiver side, the Gaussian distribution is inputted into the the same invertible neural network and recover a variable z that is similar to the loss information r with the invertible image as a dependency. this approach effectively improves the performance of image hiding, especially in dealing with loss information.
In his research, Huo proposed an image hiding architecture called FMIN \cite{huo2024fitting}. This architecture finely handles the loss information by fitting the loss information r in terms of detail content and probability distribution to improve the performance of image hiding. Although these methods process the loss information r differently, they still suffer from the problem of a large gap between the generated auxiliary variables and the loss information r. Based on these studies, this paper proposes a fully invertible image hiding method to achieve fully invertible hiding of secret images. In addition, the method is able to resist neural network-based image hiding analysis and introduces an image enhancement module, which significantly improves the robustness of the stego image. In this paper, the best results are achieved not only in terms of the quality of stego images and extracted secret images, but also in terms of resistance to image steganalysis and noise attacks.
\section{Method}
In this section, we propose a lightweight fully invertible image hiding framework, called FIIH. it skillfully hides secret images into cover images to achieve high security, strong robustness and high invisibility image hiding. Without loss of generality, we focus on the network architecture for image hiding in Sections 3.1, 3.2. our robustness module is presented in Section 3.3. our loss function is introduced in Section 3.4.

\subsection{Network architecture}
In Fig. 2 the FIIH model is shown. We first use Discrete Wavelet Transform (DWT) to decompose the secret image $X_{secret}$ into different frequency components. In this way, we can divide the image into high and low frequency domains, which enables us to effectively and quickly hide the secret image in the high frequency domain. The choice of embedding the secret information in the high-frequency domain helps to better resist the effects of noise, as demonstrated in \cite{jing2021hinet,huo2024fitting}. It is important to note that we only input the decomposed secret image $x_{secret}$ into the invertible neural network to generate the encoded secret image $s_{e}$. In order to restore back the secret image $S_{e}$ from the high and low frequency domains after the invertible neural network, we use the inverse transform of the DWT, i.e., the inverse discrete wavelet transform (IWT). Here we have two methods to hide $S_{e}$ to cover image i.e. 1)$X_{cover}$ - $S_{e}$ = $X_{stego}$, 2)$X_{cover}$ + $S_{e}$ = $X_{stego}$, to generate the stego image. Subsequently, the stego image is sent to the receiver, which is interfered by Gaussian noise, JPEG and Dropout on the way. The Field Pixel Filling Method in the figure is used only for stego images subjected to Dropout attack. After receiving the stego image, the receiver finds the corresponding cover image from the database where both the sender and receiver have the cover image. The encoded secret image $S_{e}$ is extracted from the invertible image corresponding to the above two methods i.e. 1)$X_{cover}$ - $X_{stego}$ = $S_{e}$, 2)$X_{stego}$ - $X_{cover}$ = $S_{e}$. $S_{e}$ is input into the inverse process of the invertible neural network after DWT inversion and finally the secret image is recovered. The whole process can be formulated as follows:
\begin{equation}
x_{\text {secret}}=\pi\left(X_{\text {secret}}\right)
\end{equation}
\begin{equation}
s_{\text {e}}=\mathrm{I}(x_{\text {secret}})
\end{equation}
\begin{equation}
S_{\text {e}}=\pi^{-1}\left(s_{\text {e}}\right)
\end{equation}
\begin{equation}
	X_{\text{cover}} - S_e = X_{\text{stego}} \quad \text{or} \quad X_{\text{cover}} + S_e = X_{\text{stego}}
\end{equation}
\begin{equation}
	X_{\text{cover}} - X_{\text{stego}} = S_e \quad \text{or} \quad X_{\text{stego}} - X_{\text{cover}} = S_e
\end{equation}
\begin{equation}
	s_{\text {e}}=\pi\left(S_{\text {e}}\right)
\end{equation}
\begin{equation}
	x_{\text {secret}}=\mathrm{I}^{-1}(s_{\text {e}})
\end{equation}
\begin{equation}
	X_{\text {secret}}=\pi^{-1}\left(x_{\text {secret}}\right)
\end{equation}
Where $\pi\left(* \right)$ represents our discrete wavelet transform and $\mathrm{I}\left(* \right)$ represents our image hiding model. The above two methods do not have much difference in the effectiveness of hiding single image and robustness, but method 1 is much better compared to method 2 in resisting image steganalysis. In the next section we will perform an experimental proof of steganalysis.

\begin{figure*}[!t]
\centering
\includegraphics[width=7in]{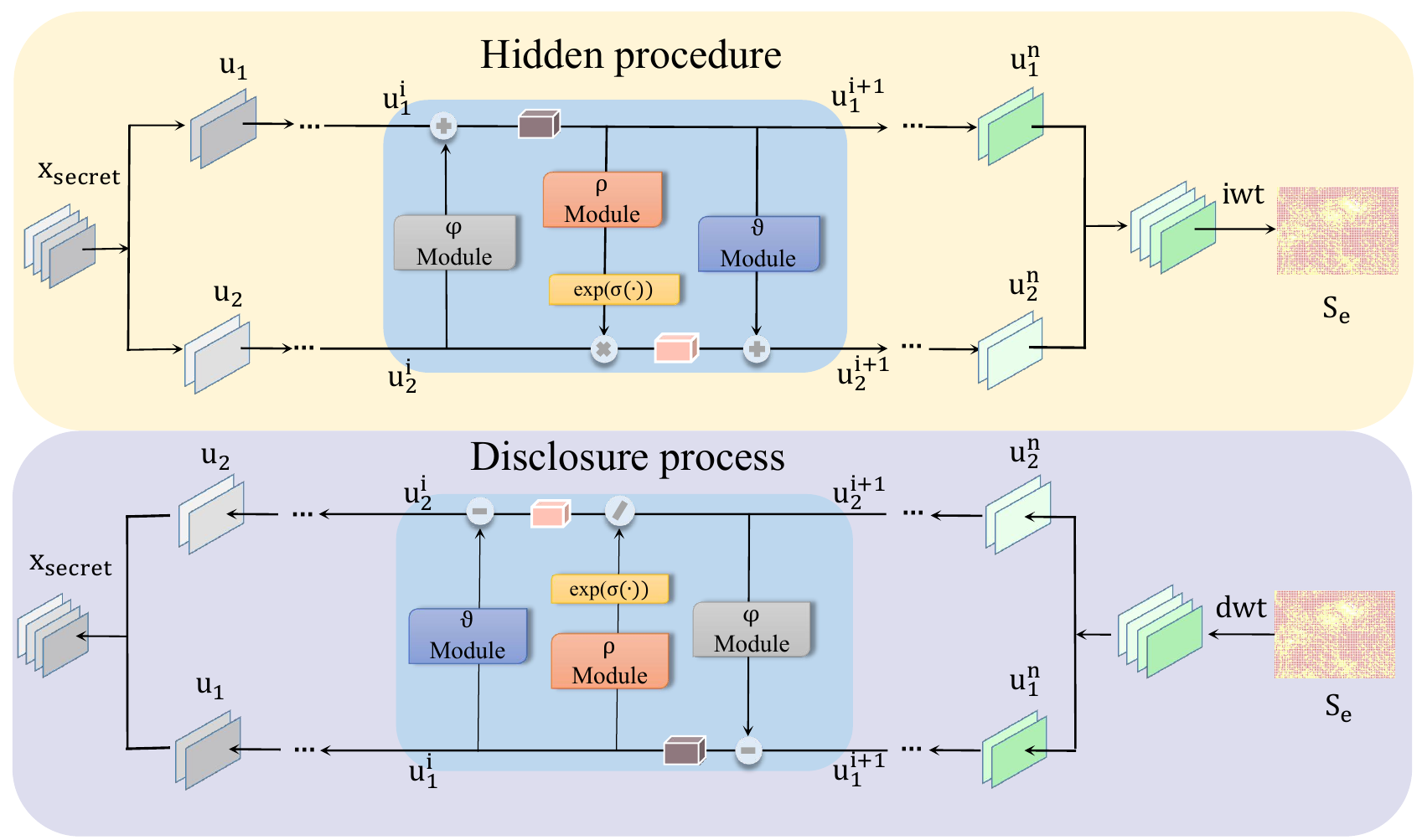} 
\caption{Architecture of invertible neural networks.}
\label{fig_3}
\end{figure*}

\subsection{Invertible neural network architecture}
Figure 3 shows our invertible neural network architecture. In the hiding process, the DWT-transformed secret image ${x}_{sceret}$ is accepted as input, and the DWT decomposes the 3*224*244 secret image into 12*112*112 wavelet subbands of LL, LH, HL, and HH. These wavelet subbands are further divided into two 6*112*112 wavelet subbands ${u}_{1}$ and ${u}_{2}$, which are then fed into a series of hidden invertible blocks. The last invertible block outputs ${u}_{1}^{\mathrm{n}}$ and ${u}_{2}^{\mathrm{n}}$, which are concatenated to generate a 3*224*244 embedded image ${S}_{e}$ by Inverse Wavelet Transform (IWT).In the revealing process, ${S}_{e}$ is extracted from the stego image, and ${S}_{e}$ is fed into the invertible neural network's inverse process, which recovers the secret image after a series of revealing invertible blocks. It is worth noting that the hidden invertible block and the revealed invertible block have the same structure and share the same network parameters; the hidden invertible block is a forward process and the revealed invertible block is an invertible process. The whole hidden invertible block contains N invertible blocks with the same architecture. The structure of the invertible block is shown in Fig. 3, and the representation of the invertible block of the hidden process is as follows:
\begin{equation}
	\mathrm{u}_{1}^{\mathrm{i}+1}=\varphi\left(\mathrm{u}_{2}^{\mathrm{i}}\right)+\mathrm{u}_{1}^{\mathrm{i}}
\end{equation}
\begin{equation}
	\mathrm{u}_{2}^{\mathrm{i}+1}=\mathrm{u}_{2}^{\mathrm{i}} * \exp \left(\sigma\left(\rho\left(\mathrm{u}_{1}^{\mathrm{i}+1}\right)\right)\right)+\vartheta\left(\mathrm{u}_{1}^{\mathrm{i}+1}\right)
\end{equation}
where $\varphi\left ( * \right )$,$\rho\left ( * \right )$,and $\vartheta\left ( * \right )$ represent arbitrary complex functions, $exp\left ( * \right )$ is an exponential function, and $\sigma$ represents a sigmoid function multiplied by a constant factor. Our functions $\varphi\left ( * \right )$,$\rho\left ( * \right )$,and $\vartheta\left ( * \right )$ use the dense blocks widely used in \cite{jing2021hinet,guan2022deepmih,zhang2019steganogan} to represent them. In the first hidden invertible block, ${u}_{1}$ represents a 6*112*112 wavelet subband of one input and ${u}_{2}$ represents a 6*112*112 wavelet subband of the other input. The representation of the invertible block that reveals the process is as follows:
\begin{equation}
	\mathrm{u}_{1}^{\mathrm{i}}=\mathrm{u}_{1}^{\mathrm{i}+1}-\varphi\left(\mathrm{u}_{2}^{\mathrm{i}+1}\right) 
\end{equation}
\begin{equation}
	\mathrm{u}_{2}^{\mathrm{i}}=\left(\mathrm{u}_{2}^{\mathrm{i}+1}-\vartheta\left(\mathrm{u}_{1}^{\mathrm{i}}\right)\right) \div \exp \left(\sigma\left(\rho\left(\mathrm{u}_{1}^{\mathrm{i}}\right)\right)\right)
\end{equation}
First, the ${S}_{e}$ obtained by the receiver is divided into two 6*112*112 wavelet subbands ${u}_{1}^{\mathrm{n}}$ and ${u}_{2}^{\mathrm{n}}$, input to in the first reveal invertible block. The last reveal invertible block outputs ${u}_{1}$ and ${u}_{2}$, concatenate the two to recover out ${X}_{secret}$.
\subsection{Robustness Module}
To enhance the robustness of the image, we introduce a noise simulation layer between the hiding and revealing networks. This noise simulation layer includes adding Gaussian noise, dropout, and simulated JPEG compression.
\subsubsection{Gaussian noise}
Gaussian noise is a statistically stochastic process that is often used to describe the uncertainty of some random variable. It is a random noise that conforms to a normal distribution. We added noise with variances of 10, 20 and 30 to the stego image for model training, aiming to improve the robustness of the stego image.
\subsubsection{Dropout}
Dropout is a random pixel point discarding operation where we randomly discard pixel points in a stego image. We gradually increase the percentage of random dropout from 0.1 to 0.9 and train the model aiming to improve the resistance of the stego image to dropout. For dropout, we propose a field pixel filling method, and Figure 4 illustrates each step. We use the four-field pixel filling method to demonstrate the processing of the receiver after receiving a stego image that has undergone a dropout.${C}_{e}^{\mathrm{1}}$ is the image received by the receiver, and we iterate through the matrix ${C}_{e}^{\mathrm{1}}$, marking the dropped pixel points as 0 to generate the mask flag matrix.

a) Equation ${C}_{e}^{\mathrm{1}}-{mask}*{X}_{cover}$ obtains the secret image matrix ${S}_{e}^{\mathrm{1}}$ after being randomly discarded.

\begin{figure}[!t]
\centering
\begin{minipage}{.5\textwidth}
  \includegraphics[width=0.3\linewidth]{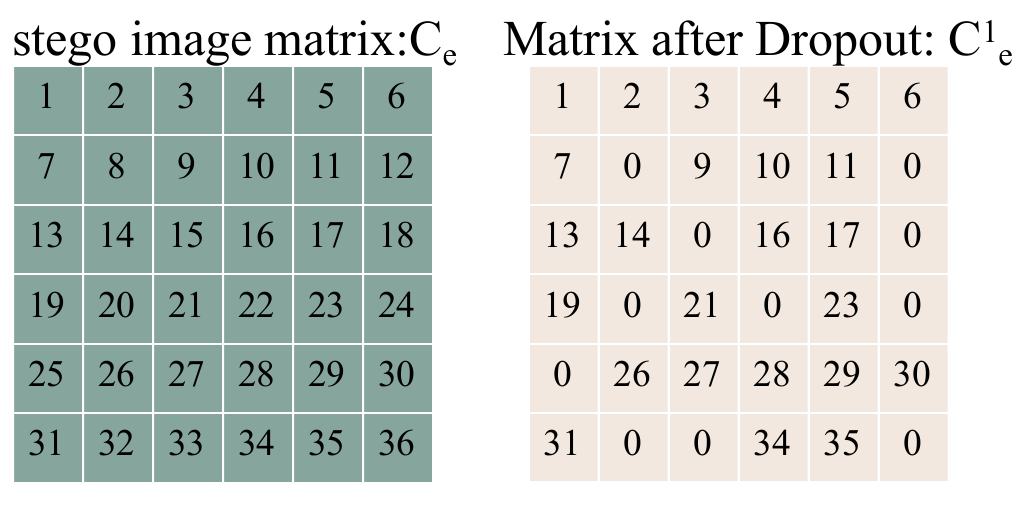}
  \includegraphics[width=0.63\linewidth]{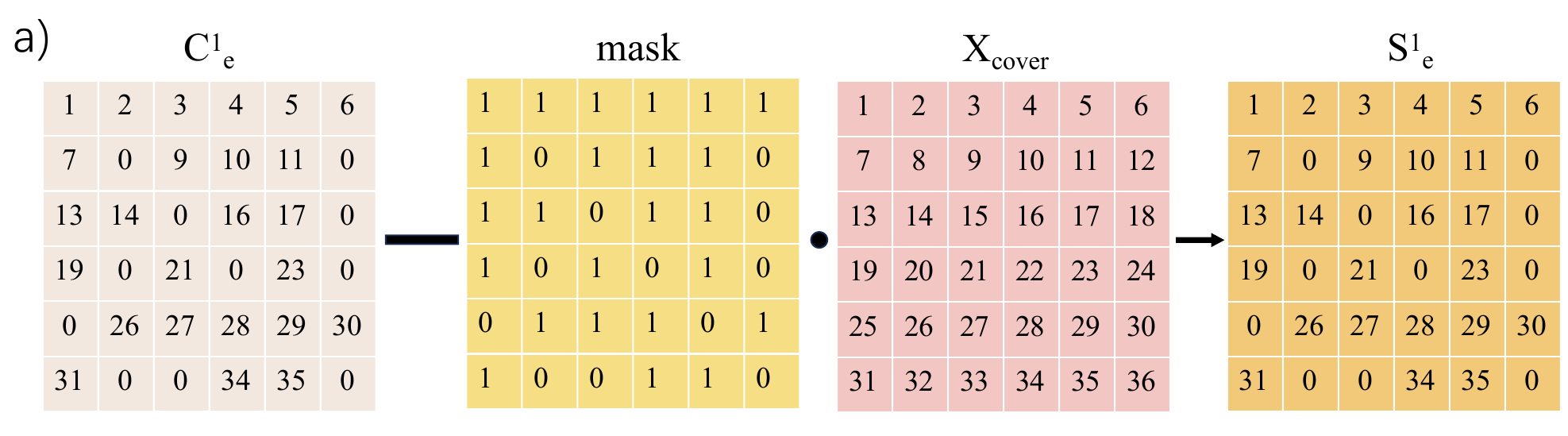}

  \includegraphics[width=0.6\linewidth]{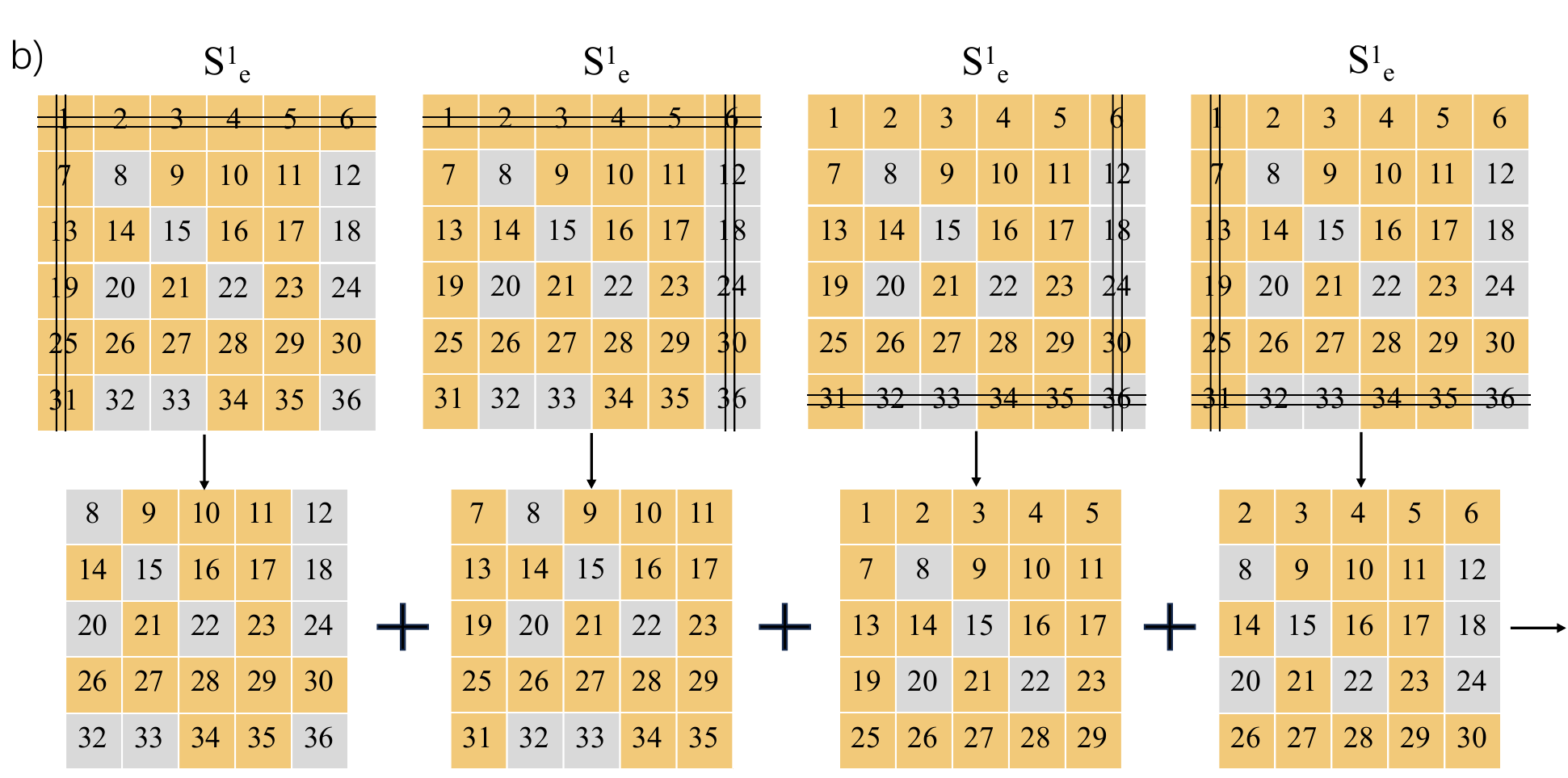}
  \includegraphics[width=0.3\linewidth]{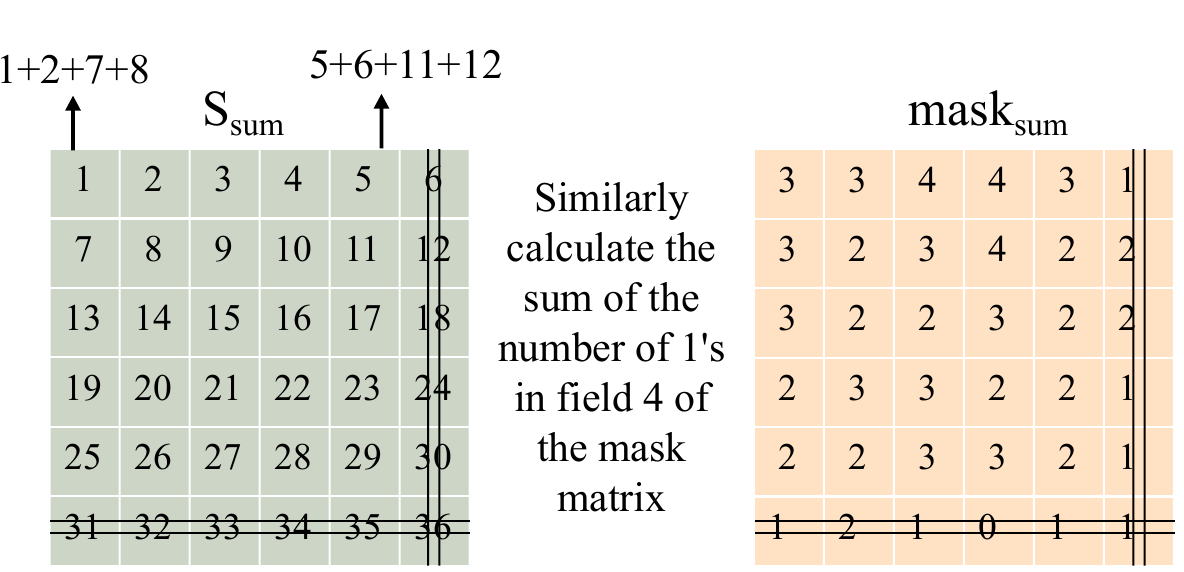}

  \includegraphics[width=0.6\linewidth]{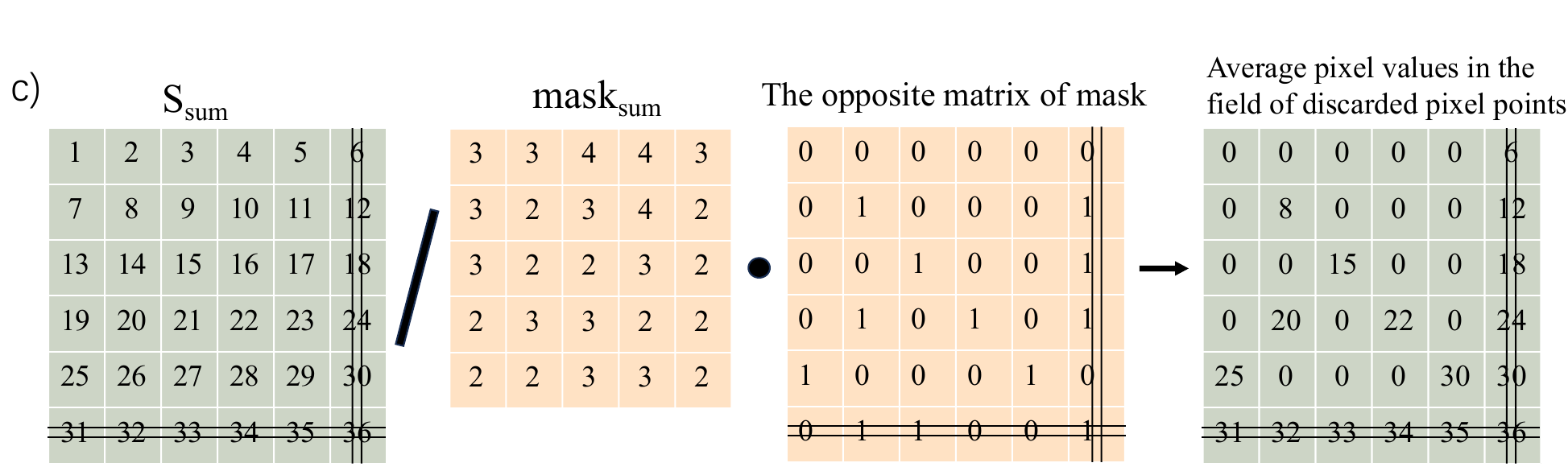}
\includegraphics[width=0.5\linewidth]{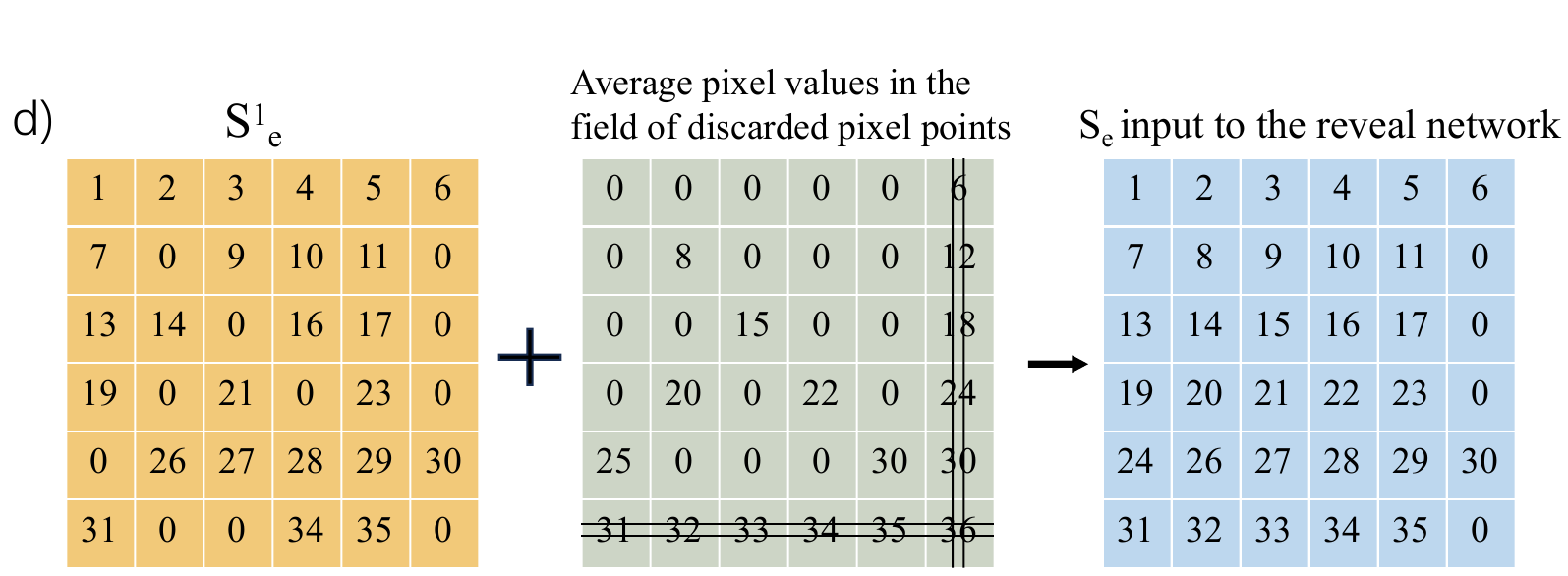}
  \caption{The process of running four-field pixel filling.}
  \label{fig_4}
\end{minipage}
\end{figure}

b) Calculate the sum of the 4 fields. The underlined rows and columns in the matrices are the removed portions, all 4 matrices are ${S}_{e}^{\mathrm{1}}$ after random discarding, and the gray points are the discarded pixel points. Summing the four matrices gives the sum of the 4 fields. It is fast and efficient to figure out the sum of the pixels of the four points, and by the same way to figure out the sum of the number of ${mask}$ matrices, the 4 fields are 1.

c) Calculate the approximate pixel values of the discarded pixel points. The average pixel value of the 4 fields was obtained by ${S}_{sum}/{mask}_{sum}*{(1-mask)}$ for replacing the discarded pixel points.

d) Restore most of the pixel points of the secret image and finally input to the reveal network to reveal the secret image.

This is the 4 Field Pixel Filling Method, in this paper we use the 9 Field Pixel Filling Method, for easy understanding and demonstration we show the 4 field method and the same for the 9 field. We recover the field of the discarded pixel point with 8 points of its field to make it smoother. In the next section, we demonstrate the Field Pixel Filling Method in action.
\subsubsection{Jpeg compression}
JPEG compression is a type of lossy compression that reduces the quality of an image by decreasing the file size, introducing a degree of distortion, and losing a certain amount of information. This type of lossy compression is very useful in scenarios such as image transfer and web page loading on the Internet. We adjusted the quality factor (QF) of JPEG compression to 20, 40 and 80, respectively, to train the model. In the training process, we first trained the model using simulated lossless JPEG compression to optimize the parameters of the model. Then, we use real lossy JPEG compression to test the anti-JPEG compression ability of stego images. We found that this training method can significantly improve the anti-JPEG compression performance of the model.
\subsection{Loss function}
The total loss function of the image hiding model consists of the hiding loss which guarantees the hiding performance, and the revealing loss which guarantees the recovery performance. 

Hiding Loss: The goal of hiding is to make the cover image and the stego image as similar as possible. We use two loss functions to ensure the hiding performance. The hiding loss is defined as follows:
\begin{equation}
	\mathrm{L}_{1}=\mathrm{MSE}=\frac{1}{\mathrm{WH}} \sum_{i=0}^{\mathrm{W}} \sum_{\mathrm{j}=0}^{\mathrm{H}}\left[\text { Stego }_{\mathrm{i},\mathrm{j}}-\operatorname{Cover}_{\mathrm{i},\mathrm{j}}\right]^{2}
\end{equation}
\begin{align}
	\mathrm{L}_{2} &= 1-\operatorname{SSIM}(\text {Stego,Cover}) \nonumber \\
	&= 1-\frac{\left(2 \mu_{\text {Stego}} \mu_{\text {Cover}}+\mathrm{c}\right)\left(2 \sigma_{\text {Stego,Cover}}+\mathrm{d}\right)}{\left(\mu_{\text {Stego}}^{2}+\mu_{\text{Cover}}^{2}+\mathrm{c}\right)\left(\sigma_{\text {Stego}}^{2}+\sigma_{\text {Cover}}^{2}+\mathrm{d}\right)}
\end{align}
Where Stego represents the stego image and Cover represents the cover image, ${L}_{1}$ calculates the average of the squares of the difference between each pixel point of the stego image and the cover image.${L}_{2}$ In order to calculate the structural similarity and luminance contrast between the stego image and the cover image, $\mu$ and $\sigma$ denote the mean and standard deviation, respectively, $\sigma_{\text {Stego,Cover}}$ represents the covariance between the stego image and the cover image, and c and d are constants.

Revealing Loss: The goal of the revealing process is to make the secret image and the extracted secret image as similar as possible. The loss function used is similar to that of the hiding process. It is defined as follows:
\begin{equation}
	\mathrm{L}_{3}=\mathrm{MSE}=\frac{1}{\mathrm{WH}} \sum_{i=0}^{\mathrm{W}} \sum_{\mathrm{j}=0}^{\mathrm{H}}\left[\text { Secret }_{\mathrm{i},\mathrm{j}}-\operatorname{Cover}_{\mathrm{i},\mathrm{j}}\right]^{2}
\end{equation}
\begin{align}
	\mathrm{L}_{4} &= 1-\operatorname{SSIM}(\text {Secret,Recovery}) \nonumber \\
	&=1-\frac{\left(2 \mu_{\text{Secret}}\mu_{\text {Recovery}}+\mathrm{c}\right)\left(2 \sigma_{\text {Secret,Recovery}}+\mathrm{d}\right)}{\left(\mu_{\text {Secret}}^{2}+\mu_{\text {Recovery}}^{2}+\mathrm{c}\right)\left(\sigma_{\text {Secret}}^{2}+\sigma_{\text{Recovery}}^{2}+\mathrm{d}\right)}
\end{align}
where secret represents the secret image embedded in the cover image,and Recovery represents the secret image recovered from the stego image.

Total loss function of the image hiding model:The total loss function ${L}_{total }$ is the weighted sum of the hiding losses ${~L}_{1}$ and ${~L}_{2}$,and the revealing losses ${~L}_{3}$ and ${~L}_{4}$.The specific formula is as follows:
\begin{equation}
	\mathrm{L}_{\text {total }}=\lambda_{1}\mathrm{~L}_{1}+\lambda_{2}\mathrm{~L}_{2}+\lambda_{3}\mathrm{~L}_{3}+\lambda_{4}\mathrm{~L}_{4}
\end{equation}
where $\lambda_{1},\lambda_{2},\lambda_{3},\lambda_{4}$ balance the weights of different loss terms. During training,$\lambda_{1},\lambda_{2},\lambda_{3},\lambda_{4}$ are taken as 50, 50, 1, and 1 respectively.
\section{Experiment}
In this section, we evaluate the performance of the proposed FIIH framework. Section 4.1 details our experimental setup. To the best of our knowledge, there has not been any method to achieve fully invertible image hiding. In Section 4.2, we will compare the proposed FIIH method with other state-of-the-art methods in the case of hiding a single image. Section 4.3 will focus on evaluating the comparison of FIIH with other state-of-the-art methods in terms of robustness of stego images. In Section 4.4, we will perform ablation experiments to demonstrate the effectiveness of the method proposed in this paper. Finally, in Section 4.5, we will perform a security analysis and compare it with other state-of-the-art methods.
\subsection{Experimental Settings}
\subsubsection{Data setting}
For network training, we used the COCO dataset to train our FIIH architecture. We selected 50,000 images from the COCO dataset and cropped them randomly at 224 × 224. The test dataset consists of 1000 images from the DIV2K \cite{agustsson2017ntire}, ImageNet \cite{russakovsky2015imagenet}, and Paris StreetView datasets, which were also subjected to 224 × 224 random cropping. It is worth noting that we chose these three datasets as our test datasets, thus fully validating the generalization ability of our model. The number of hidden and displayed blocks was set to 16, and the image hiding model was performed for 130 iterations.The small batch size was set to 8. Half of them were used as cover images while the rest were used as secret images. We used the Adam \cite{kingma2014adam} optimizer with an initial learning rate of 0.0001. The learning rate was halved every 30 iterations and the weights were updated every 1 batch of training.

\renewcommand{\floatpagefraction}{.9}
\begin{table*}[tbp]
    \center
    {\small
        \caption{Our approach is compared on four datasets (DIVK, COCO, ImageNet, Paris StreetView) with HIWI, Hinet, and FMIN.}
        \begin{tabularx}{\textwidth}{l|*{8}{X}}
            \hline
            \hline
            \multirow{3}*{Methods} & \multicolumn{8}{c}{Cover/Stego image pair} \\
            \cline{2-9}
            & \multicolumn{2}{c}{DIV2K} & \multicolumn{2}{c}{COCO} & \multicolumn{2}{c}{ImageNet} & \multicolumn{2}{c}{Paris StreetView} \\
            \cline{2-9}
            & PSNR(dB) & SSIM & PSNR(dB) & SSIM & PSNR(dB) & SSIM & PSNR(dB) & SSIM \\
            \hline
            HIWI[40] & 37.04 & 0.9845 & 36.45 & 0.9489 & 36.02 & 0.9465 & 36.80 & 0.9861 \\
            HiNet & 46.53 & 0.9932 & 41.21 & 0.9852 & 41.60 & 0.9858 & 46.64 & 0.9936 \\
            \underline{FMIN} & \underline{48.45} & \underline{0.9957} & \underline{42.85} & \underline{0.9901} & \underline{42.92} & \underline{0.9905} & \underline{48.95} & \underline{0.9950} \\
            \textbf{ours} & \textbf{51.08} & \textbf{0.9966} & \textbf{47.23} & \textbf{0.9922} & \textbf{48.89} & \textbf{0.9947} & \textbf{51.57} & \textbf{0.9965} \\
            \hline
            \hline
            \multirow{3}*{Methods} & \multicolumn{8}{c}{Secret/Recovery image pair} \\
            \cline{2-9}
            & \multicolumn{2}{c}{DIV2K} & \multicolumn{2}{c}{COCO} & \multicolumn{2}{c}{ImageNet} & \multicolumn{2}{c}{Paris StreetView} \\
            \cline{2-9}
            & PSNR(dB) & SSIM & PSNR(dB) & SSIM & PSNR(dB) & SSIM & PSNR(dB) & SSIM \\
            \hline
            HIWI[40] & 39.21 & 0.9831 & 33.02 & 0.9376 & 32.75 & 0.933 & 39.03 & 0.9845 \\
            HiNet & 46.24 & 0.9924 & 40.10 & 0.9846 & 40.03 & 0.9807 & 46.04 & 0.9926 \\
            \underline{FMIN} & \underline{48.29} & \underline{0.9954} & \underline{42.76} & \underline{0.9899} & \underline{42.71} & \underline{0.9887} & \underline{48.50} & \underline{0.9952} \\
            \textbf{ours} & \textbf{51.47} & \textbf{0.9969} & \textbf{51.16} & \textbf{0.9975} & \textbf{51.02} & \textbf{0.9969} & \textbf{51.32} & \textbf{0.9966} \\
            \hline
    \end{tabularx}}
\end{table*}

\begin{figure*}[!t]
\centering
\includegraphics[width=7in]{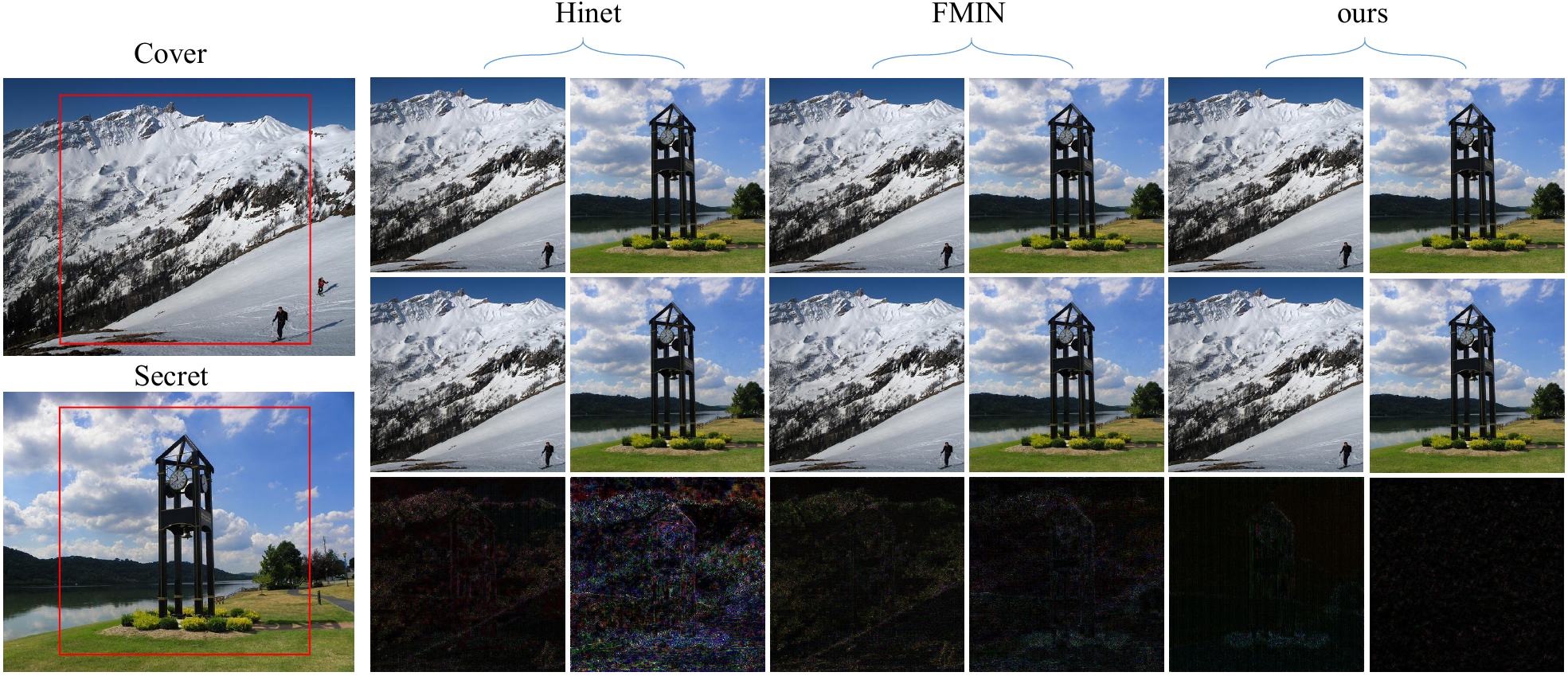} 
\caption{Visual comparison of Hinet, FMIN and our method on single image hiding, the first row of the right part is the cover image and the secret image from left to right, the second row is the loaded image and the secret image extracted from the loaded image, and the third row is the residual image *20.}
\label{fig_5}
\end{figure*}

\begin{figure}[!t]
\centering
\begin{minipage}{.5\textwidth}
  \centering
  \includegraphics[width=0.45\linewidth]{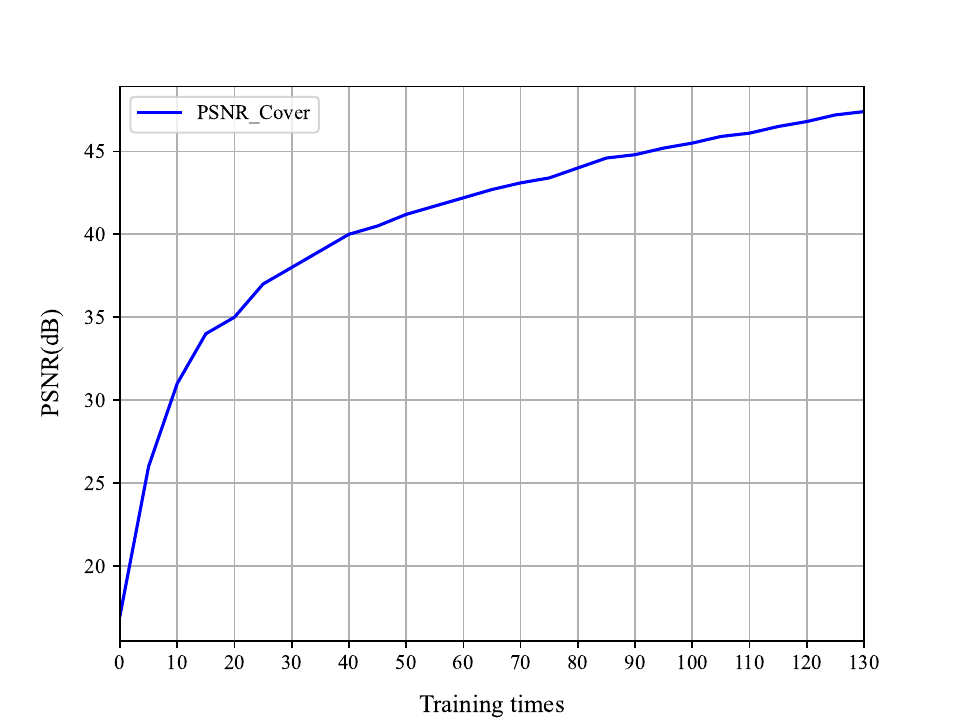}
  \includegraphics[width=0.45\linewidth]{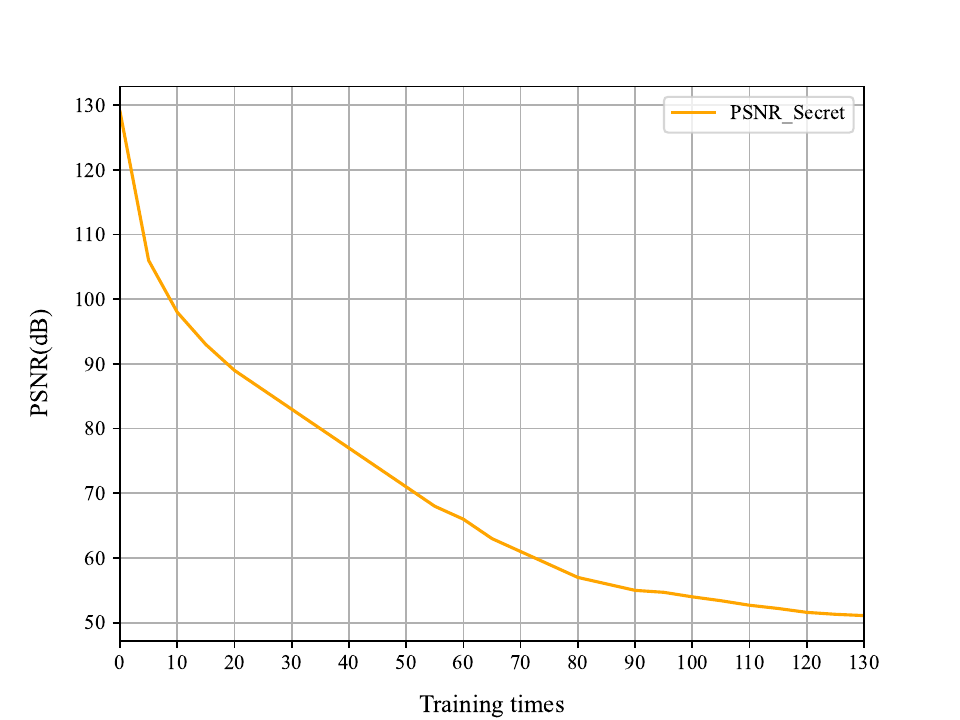}
  \caption{The training process of our FIIH on cover/stego image and secret/recovery image pairs, with the cover/stego image on the left and the secret/recovery image on the right.}
  \label{fig_6}
\end{minipage}
\end{figure}

\subsubsection{Evaluation index}
In order to evaluate the hiding and revealing performance of FIIH, we use two metrics to evaluate the quality of cover/stego image and secret/recovery image pairs, including Peak Signal-to-Noise Ratio (PSNR), Structural Similarity Index (SSIM).

PSNR is a measure of image or video quality, which is used to assess the similarity between the original image and the compressed or processed image.The formula for PSNR is as follows:
\begin{equation}
	\operatorname{PSNR}=10 * \log _{10} \frac{\mathrm{MAX}^{2}}{\mathrm{MSE}}
\end{equation}
where MAX represents the maximum pixel value of the image,and MSE denotes the mean squared error,which indicates the difference between the original image and the processed image.

SSIM is a metric that evaluates the quality of an image or video in terms of structural similarity and luminance contrast between two images.The formula for SSIM is as follows:
\begin{equation}
	\operatorname{SSIM}(x,y)=\frac{\left(2 \mu_{x} \mu_{y}+c\right)\left(2 \sigma_{x y}+d\right)}{\left(\mu_{x}^{2}+\mu_{y}^{2}+c\right)\left(\sigma_{x}^{2}+\sigma_{y}^{2}+d\right)}
\end{equation}
where x and y represent the original image and the processed image,respectively.$\mu$ and $\sigma$ denote the mean and standard deviation,$\sigma_{x y}$ and  represents the covariance between x and y.c and d are constants.

\begin{table*}[tbp]
    \center
    {\small
        \caption{For Gaussian noise, our method is compared with Hinet, RIIS and CRoSS on the DIVK and COCO datasets.}
        \begin{tabularx}{\textwidth}{l|*{8}{X}}
            \hline
            \hline
            \multirow{3}*{Methods} & \multicolumn{4}{c}{DIV2K} & \multicolumn{4}{c}{COCO} \\
            \cline{2-9}
            & Clean & $\sigma = 10$ & $\sigma = 20$ & $\sigma = 30$ & Clean & $\sigma = 10$ & $\sigma = 20$ & $\sigma = 30$ \\
            \cline{2-9}
            & PSNR(dB) & PSNR(dB) & PSNR(dB) & PSNR(dB) & PSNR(dB) & PSNR(dB) & PSNR(dB) & PSNR(dB) \\
            \hline
            HiNet & \underline{46.24} & 12.91 & 11.54 & 10.23 & \underline{46.24} & 12.73 & 11.23 & 10.12 \\
            RIIS & 43.78 & \underline{26.03} & 18.89 & 15.85 & 43.78 & \underline{25.32} & 17.92 & 15.46 \\
            CRoSS & 23.79 & 21.89 & \underline{20.19} & \underline{18.77} & 23.79 & 21.89 & \underline{20.19} & \underline{18.77} \\
            \textbf{ours} & \textbf{51.47} & \textbf{28.06} & \textbf{23.02} & \textbf{21.68} & \textbf{51.46} & \textbf{27.10} & \textbf{22.30} & \textbf{21.12} \\
            \hline
            \hline
    \end{tabularx}}
\end{table*}

\begin{figure*}[!t]
\centering
\includegraphics[width=7in]{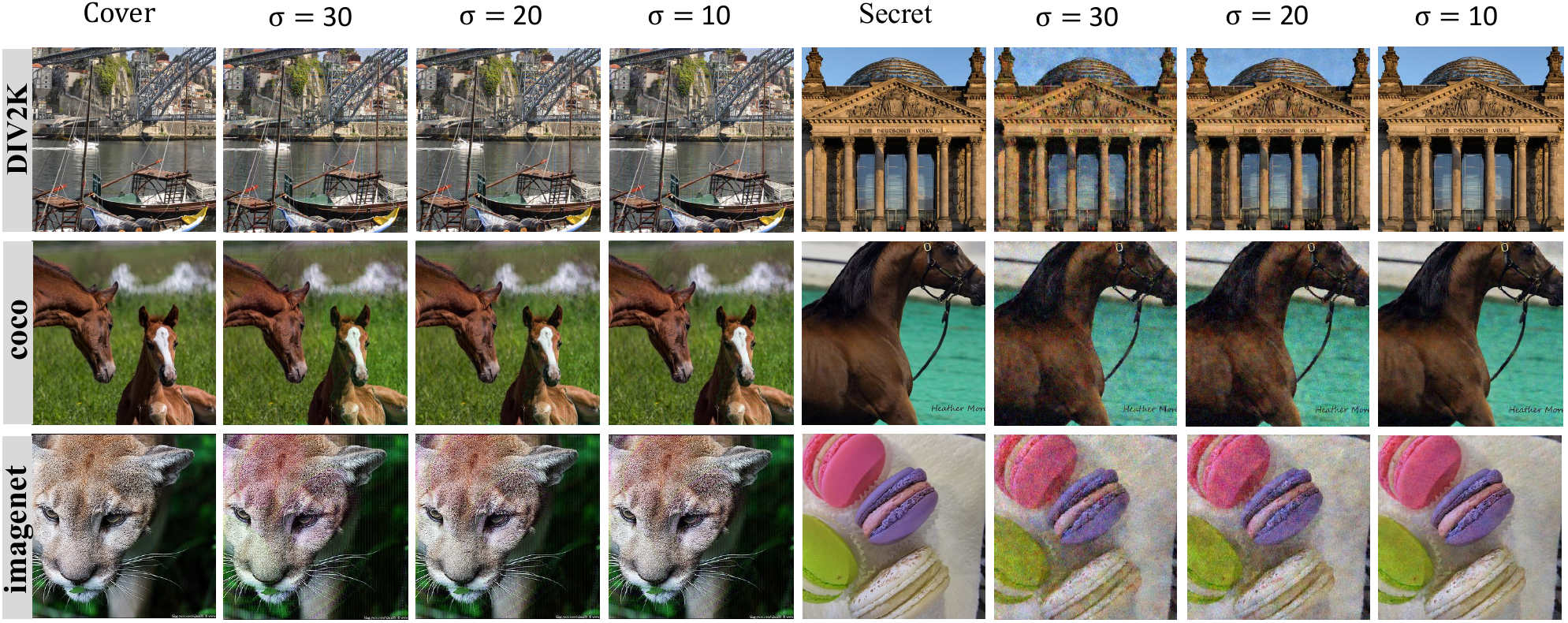} 
\caption{Our method is tested on DIVK, COCO and imagenet datasets. The first column is the cover image, the second, third and fourth columns are the loaded image under different Gaussian noise interference, the fifth column is the secret image, and the sixth, seventh and eighth columns are the secret image extracted under different Gaussian noise interference.}
\label{fig_7}
\end{figure*}

\subsection{Single Image Hide}
\textbf{Quantitative results}. To assess the excellence of our approach, Table 1 compares the numerical results of our FIIH with HiNet\cite{jing2021hinet}, HIWI\cite{baluja2019hiding}, and FMIN \cite{huo2024fitting}.HIWI uses one network to handle the hiding process and another to handle the revealing process, and fails to achieve the reversibility of both the hiding network and the revealing network, which leads to significant data loss. The superiority of invertible neural networks has been demonstrated in HiNet, ISN \cite{lu2021large}, and DeepMIH \cite{guan2022deepmih}, i.e., invertible neural networks are able to realize the reversibility of both hidden and revealed networks, which share the same parameters. However, when dealing with loss information r, they choose to discard the loss information r and use Gaussian distribution or constant matrix as auxiliary variable z, which leads to loss of information and fails to achieve reversibility of data. Although FMIN and RIIS process the loss information r to some extent, they still fail to realize the invertible extraction of data.

Our FIIH architecture realizes a fully invertible image hiding process. Among the current single-image hiding, FMIN and HiNet are leading in terms of image quality. From Table 1, bold font indicates our results and underline font indicates sub-optimal results. It can be clearly seen that our FIIH significantly outperforms the other methods on all metrics for cover/stego image pairs and secret/recovery image pairs on the DIV2K, coco, ImageNet, and Paris StreetView datasets. Specifically, for cover/stego image pairs, our FIIH improves 2.63 dB, 4.38 dB, 5.97 dB, and 2.62 dB on PSNR over blue font on these four datasets, respectively.For secret/recovery image pairs, we improve 3.18 dB, 8.40 dB, 8.31 dB, and 2.82 dB on PSNR.In addition to the PSNR, there is also an improvement in SSIM, although the numerical improvement is not very significant due to the fact that SSIM is close to one. The experiments demonstrate that for other methods, the choice of dataset has a significant impact on the image quality, whereas for our method it is essentially unaffected. Our experimental results reach the best level so far.

\textbf{Qualitative results.} Figure 5 shows a visual comparison between FIIH with HiNet and FMIN. In order to highlight the difference between the original and generated images, we magnified the pixel-level error by a factor of 20. We observe that the errors of both our generated stego images and revealed secret images are smaller than those generated by HiNet and FMIN, which is consistent with the objective comparison results. We observe that HiNet and FMIN, which do not address invertible hiding of data, have information differences in the residual maps between the secret image and the extracted secret image. Whereas, our proposed fully invertible neural network approach extracts the secret image almost perfectly with reduced information loss.

\begin{table*}[tbp]
    \center
    {\small
        \caption{For dropout, our method is compared with Hinet and FMIN on the DIV2K and COCO datasets.}
        \begin{tabularx}{\textwidth}{l|*{10}{X}}
            \hline
            \hline
            \multirow{3}*{Methods} & \multicolumn{5}{c}{DIV2K} & \multicolumn{5}{c}{COCO} \\
            \cline{2-11}
            & 0.1 & 0.3 & 0.5 & 0.7 & 0.9 & 0.1 & 0.3 & 0.5 & 0.7 & 0.9 \\
            \cline{2-11}
            & PSNR(dB) & PSNR(dB) & PSNR(dB) & PSNR(dB) & PSNR(dB) & PSNR(dB) & PSNR(dB) & PSNR(dB) & PSNR(dB) & PSNR(dB) \\
            \hline
            HiNet & \underline{35.81} & \underline{31.62} & \underline{29.63} & \underline{26.12} & \underline{24.31} & \underline{32.13} & \underline{28.24} & \underline{26.52} & \underline{24.27} & \underline{21.12} \\
            FMIN & 35.47 & 30.52 & 28.34 & 25.23 & 21.76 & 31.87 & 27.64 & 26.23 & 23.82 & 20.35 \\
            \textbf{ours} & \textbf{41.04} & \textbf{34.41} & \textbf{32.24} & \textbf{28.38} & \textbf{24.74} & \textbf{37.78} & \textbf{32.25} & \textbf{29.06} & \textbf{25.47} & \textbf{22.13} \\
            \hline
            \hline
    \end{tabularx}}
\end{table*}

\begin{figure*}[!t]
\centering
\includegraphics[width=7in]{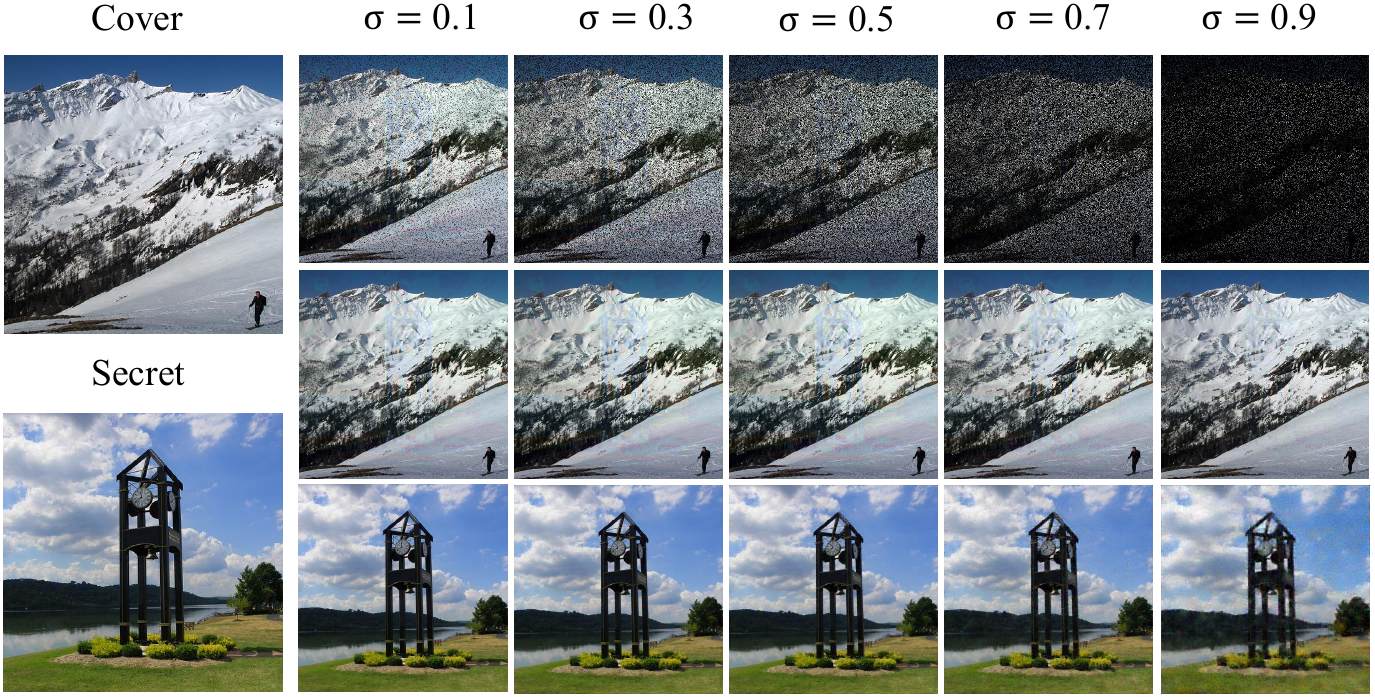} 
\caption{The first line is the loaded image at different dropout ratios, the second line is the loaded image recovered by the field filling method, and the third line is the extracted secret image.}
\label{fig_8}
\end{figure*}

\textbf{Convergence Experiments.} Compared to HiNet and FMIN, our FIIH shows a significant advantage in convergence speed. Under the same dataset conditions, HiNet requires more than two thousand cycles to achieve optimal results, while FMIN requires about 500 cycles. In contrast, our method requires only 130 cycles to complete training, as shown in Fig. 6. Our method extracts secret images with a peak signal-to-noise ratio of 110 at the beginning of training, demonstrating that we have achieved a fully invertible image hiding method.
As the number of training sessions increases, to improve the quality of the stego image, our model discards a small portion of the details. When the PSNR of the stego image reaches 48, the quality of the extracted secret image still reaches 51+. This result is the best so far as I know. Of course, we can also reduce the quality of the stego image to 40, at which point the PSNR of the extracted secret image can be as high as an astonishing 80. Experiments have shown that our method significantly reduces the training time, making it a fast, lightweight, and optimally effective method for image hiding.
\subsection{Robustness experiments}
\subsubsection{Gaussian noise}
We analyzed both quantitative and qualitative results.

\textbf{Quantitative results.} In order to verify whether our method can resist Gaussian noise, we adjust the variance of Gaussian noise, which is set to 10, 20, and 30, respectively.To evaluate the superiority of our method in resisting Gaussian noise, we compare it with Hinet, CRoSS, and RIIS, of which CRoSS and RIIS are the current optimal results. Table 2 shows the numerical results of the comparison.Hinet and RIIS use invertible neural network for image hiding.CRoSS uses diffusion model for image hiding and is the first paper to use diffusion model in image hiding. As can be seen in Table 2, our FIIH significantly outperforms all methods compared on both datasets for all metrics of secret/recovery image pairs. The underline font represents the suboptimal results and the bold font represents the results of our proposed method. Specifically, at $\sigma$ = 10, for secret/recovery image pairs, our FIIH outperforms the suboptimal effect by 6-7 dB per dataset on PSNR. at $\sigma$ = 20, 30, for secret/recovery image pairs, our FIIH outperforms the suboptimal effect by 2-3 dB per dataset on PSNR.

\textbf{Qualitative results.} The visualization of our FIIH with Gaussian noise variance of 10, 20, and 30 is shown in Fig. 7. We trained the model using the coco dataset, and Fig. 7 shows the results for two other datasets, illustrating the good generalization ability of our method. It is observed that the quality of both the stego image and the extracted secret image improves as the variance of the Gaussian noise decreases. At a variance of 30, the content of the extracted secret image can still be clearly seen, indicating that our method is able to resist Gaussian noise to some extent.

\begin{table*}[tbp]
    \center
    {\small
        \caption{For Jpeg compression, our method is compared with Hinet, RIIS, and CRoSS on the DIV2K and COCO datasets.}
        \begin{tabularx}{\textwidth}{l|*{8}{X}}
            \hline
            \hline
            \multirow{3}*{Methods} & \multicolumn{4}{c}{DIV2K} & \multicolumn{4}{c}{COCO} \\
            \cline{2-9}
            & Clean & QF=20 & QF=40 & QF=80 & Clean & QF=20 & QF=40 & QF=80 \\
            \cline{2-9}
            & PSNR(dB) & PSNR(dB) & PSNR(dB) & PSNR(dB) & PSNR(dB) & PSNR(dB) & PSNR(dB) & PSNR(dB) \\
            \hline
            HiNet & \underline{46.24} & 7.03 & 9.78 & 11.54 & \underline{46.24} & 6.92 & 9.54 & 11.23 \\
            RIIS & 43.78 & \underline{22.03} & \underline{27.12} & \underline{28.25} & 43.78 & \underline{22.76} & \underline{24.75} & \underline{25.87} \\
            CRoSS & 23.79 & 21.74 & 22.74 & 23.51 & 23.79 & 21.74 & 22.74 & 23.51 \\
            \textbf{ours} & \textbf{51.47} & \textbf{24.15} & \textbf{27.42} & \textbf{31.21} & \textbf{51.12} & \textbf{23.25} & \textbf{25.37} & \textbf{29.17} \\
            \hline
            \hline
    \end{tabularx}}
\end{table*}

\begin{figure*}[!t]
\centering
\includegraphics[width=7in]{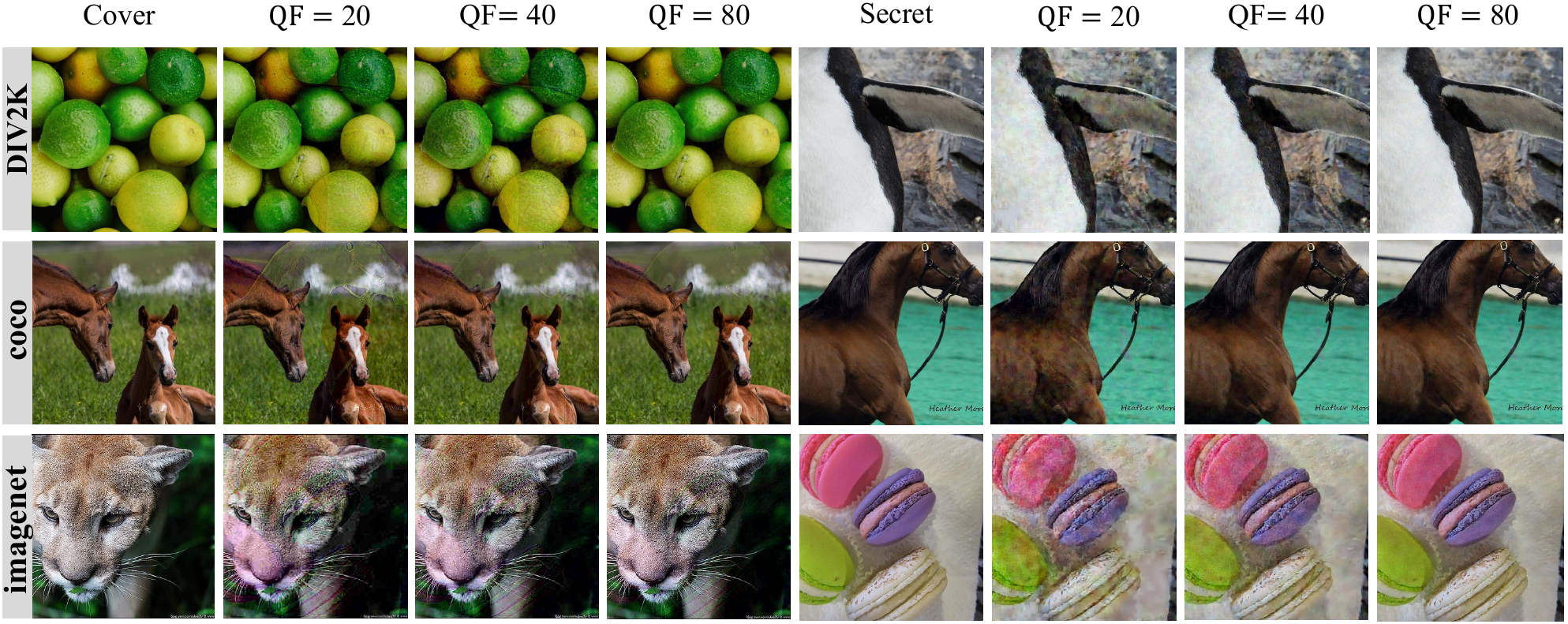} 
\caption{Our method is tested on DIVK, COCO and imagenet datasets. The first column is the cover image, the second, third and fourth columns are the loaded image under different JPEG compression factors, the fifth column is the secret image, and the sixth, seventh and eighth columns are the secret image extracted under different JPEG compression factors.}
\label{fig_9}
\end{figure*}

\subsubsection{Dropout}
We analyzed both quantitative and qualitative results.

\textbf{Quantitative Results.} To verify the ability of our method to resist pixel dropout, we randomly discard pixel points of stego images according to percentage, with $\sigma$ set to 0.1, 0.3, 0.5, 0.7, and 0.9, respectively.In order to evaluate the excellence of our method in resisting dropout, FIIH is compared with Hinet and FMIN. The numerical results of the comparison are demonstrated in Table 3. As can be seen from Table 3, our FIIH significantly outperforms all compared methods on all metrics for secret/recovery image pairs on both datasets. The underline font represents the suboptimal effect and the bold font represents the effect of our proposed method. Specifically, at $\sigma$ = 0.1, for secret/recovery image pairs, our FIIH outperforms the suboptimal effect by 5-6 dB in PSNR on each dataset.At $\sigma$ = 0.3, our FIIH outperforms the suboptimal effect by 3-4 dB in PSNR on each dataset.At $\sigma$ = 0.5, 0.7, our FIIH outperforms the suboptimal effect by 2-3 dB in PSNR on each dataset. At $\sigma$ = 0.9, our FIIH improved by 0.4-1 dB over the suboptimal effect for each dataset on the PSNR.

\textbf{Qualitative results.} The visualization of our FIIH method on the coco dataset when the pixel points are discarded is shown in Fig. 8. As $\sigma$ increases, the content of the stego image is gradually unrecognizable. After the Field Pixel Filling Method, we are still able to extract the secret image accurately even at $\sigma$ = 0.9. We will prove its effectiveness in the next section.
\subsubsection{Jpeg compression}
We analyzed both quantitative and qualitative results.

\textbf{Quantitative Results.}To verify whether our FIIH is able to resist JPEG compression, we performed JPEG compression on stego images with QFs set to 20, 40, and 80, respectively.To evaluate the excellence of FIIH in resisting JPEG compression, FIIH was compared with Hinet, CRoSS, and RIIS, where CRoSS and RIIS are the current methods with the best results. The numerical results of the comparison are demonstrated in Table 4. As can be seen from Table 4, our FIIH significantly outperforms all compared methods on all metrics for secret/recovery image pairs on both datasets. The underline font represents the suboptimal effect and the bold font represents the effect of our proposed method. Specifically, FIIH outperforms the suboptimal effect by 3-4 dB in each dataset at QF = 80. at QF = 40, FIIH outperforms the suboptimal effect by 0.3-1 dB in each dataset at QF = 40. at QF = 20, FIIH outperforms the suboptimal effect by 1-2 dB in each dataset at PSNR.

\textbf{Qualitative results.} The visualization of our FIIH at QFs of 20, 40, and 80 is shown in Fig. 9. We observe that after JPEG compression, especially on the coco dataset, the stego image appears as an artifact of the secret image. For this situation, we consider two solutions. One is to improve the quality of the stego image by reducing the quality of the extracted secret image. The second is to first disrupt the pixel points of the secret image in a regular manner and then hide them in the cover image, and the receiver restores the secret image according to the regularity. When the QF is 20, the content of the extracted secret image can still be clearly seen, which shows that our method has some resistance to JPEG compression.

\subsection{Ablation Experiments}
\subsubsection{Effectiveness of Innovation Architecture}
To verify the effectiveness of the innovative architecture proposed in this paper, the difference between the innovative architecture and the traditional method is shown in Fig. 1. Our method only inputs the secret image into the invertible neural network with no loss of information and realizes a fully invertible image hiding method. Experimental comparisons were conducted and the results are shown in Table 5. Specifically, from the first and third rows of Table 5, it can be seen that for single-image hiding, the use of the innovative architecture improves the PSRN by about 6 dB on cover/stego image pairs and about 11 dB on secret/recovery image pairs.It is well illustrated that the innovative architecture is more effective as compared to the traditional methods. The most important thing about our innovative architecture compared to traditional neural network methods is that it realizes invertible hiding of data, which fully illustrates the importance of realizing invertible hiding of data.
\subsubsection{Effectiveness of Invertible Neural Networks}
To verify the effectiveness of the invertible neural network, we conducted experiments by replacing the invertible neural network in Fig. 1 with a convolutional neural network and compared the results as shown in Table 5. From the second and third rows of Table 5, it can be seen that for single-image hiding, the use of invertible neural network improves the PSNR by about 10 dB on cover/stego image pairs and secret/recovery image pairs.This is a good illustration of the greater efficiency of the invertible neural network, which is the key to the realization of both the hiding network and the revealing network.

\begin{table}[htbp]
    \centering
    \caption{Validating the effectiveness of our proposed innovative architecture and invertible neural network on single graph hiding, the third row is our FIIH.}
    \begin{tabularx}{\linewidth}{cccccc}
        \toprule
        \multirow{2}{*}{\makecell{Innovation \\ Architecture}} & \multirow{2}{*}{\makecell{Invertible \\ Neural Networks}} & \multicolumn{2}{c}{Stego}  & \multicolumn{2}{c}{Recovery} \\
        \cmidrule(lr){3-6}
         &  & PSRN & SSIM & PSRN & SSIM \\
        \midrule
        $\times$ & $\checkmark$ & 41.21 & 0.9852 & 40.10 & 0.9835 \\
        $\checkmark$ & $\times$ & 39.18 & 0.9834 & 39.13 & 0.9825 \\
        $\checkmark$ & $\checkmark$ & 47.23 & 0.9922 & 51.16 & 0.9975 \\
        \bottomrule
    \end{tabularx}
\end{table}

\subsubsection{Effectiveness of domain complementarity laws}
In order to verify the effectiveness of the Field Pixel Filling Method, we compare no Field Pixel Filling Method, 4 Field Pixel Filling Method and 9 Field Pixel Filling Method, and the results are shown in Table 6. From the first and third rows of Table 6, it can be seen that using the Field Pixel Filling Method improves the PSRN by 3dB on cover/stego image pairs and about 1dB on secret/recovery image pairs when fighting against random dropout.This fully illustrates the effectiveness of the Field Pixel Filling Method. And 9 Field Pixel Filling Method is more effective than 4 Field Pixel Filling Method. Our proposed Field Pixel Filling Method only operates on matrices and has no loop traversal operation, which does not affect the running time.

\begin{table}[htbp]
    \centering
    \caption{Verify the validity of our proposed Field Pixel Filling Method on a random dropout, with the third line being our FIIH.}
    \begin{tabularx}{\linewidth}{ccccccc}
        \toprule
        \multirow{2}{*}{\makecell{4 Field \\ Pixel Filling}} & \multirow{2}{*}{\makecell{9 Field \\ Pixel Filling}} & \multicolumn{5}{c}{Noise Level ($\sigma$)} \\
        \cmidrule(lr){3-7}
         &  & 0.1 & 0.3 & 0.5 & 0.7 & 0.9 \\
        \midrule
        $\times$ & $\times$ & 35.27 & 29.40 & 27.34 & 25.27 & 22.23 \\
        $\checkmark$ & $\times$ & 36.12 & 30.87 & 28.27 & 26.21 & 22.85 \\
        $\times$ & $\checkmark$ & 41.04 & 34.41 & 32.24 & 28.38 & 24.74 \\
        \bottomrule
    \end{tabularx}
\end{table}

\subsection{Security Analysis}
Steganalysis is an important part of the image hiding task to measure the security of the target image. Specifically, an analysis tool is used to measure the distinction between images with embedded secret information and images without embedded secret information. Image steganalysis tools include traditional statistical methods and deep learning based image steganalysis.

For traditional statistical methods are basically unable to detect deep learning based image steganography, it has been shown in \cite{jing2021hinet,guan2022deepmih} that the accuracy of traditional statistical methods for detecting image steganography based on invertible neural networks is 0.5, which is equivalent to random guessing.

For deep learning based image steganalysis methods we use SRNet \cite{boroumand2018deep} network. We use FIIH to generate stego images as a dataset for training SRNet. The experimental results are shown in Fig. 10, where FIIH is the scheme taken in this paper: $X_{cover}$ - $S_{e}$ = $X_{stego}$, and FIIH+ is the change from $X_{cover}$ - $S_{e}$ = $X_{stego}$ to $X_{cover}$ + $S_{e}$ = $X_{stego}$. we can observe that both FIIH and FIIH+ improve a lot more than all the other methods to achieve a very good result.FIIH improves again a lot more than FIIH+, and in FIIH it is the change of the cover image by subtracting part of the information from the cover image and actually not embedding the secret image into the cover image.FIIH+ is adding $S_{e}$ to the cover image and embedding $S_{e}$ into the cover image.As the training dataset increases, the detection success rate rises more significantly.

\begin{figure}[!t]
\centering
\includegraphics[width=3.5in]{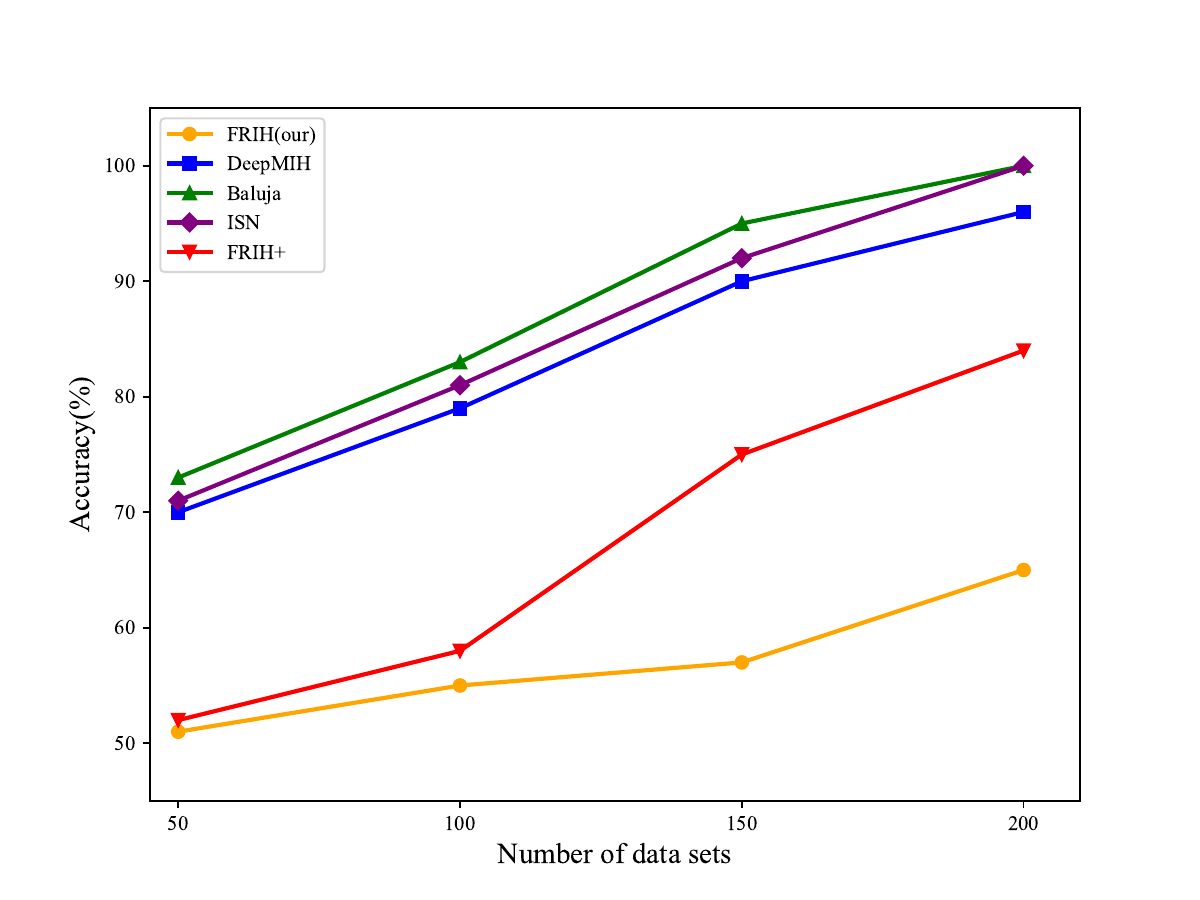}
\caption{The ability of different methods to resist SRnet analysis. Note that the closer the precision is to 50\%, the higher the resistance to steganalysis it can achieve.}
\label{fig_10}
\end{figure}

\section{Conclusion and future work}
We observe that previous image hiding methods based on invertible neural networks only achieve reversibility of the network and do not achieve invertible hiding on the data. In this paper, we propose for the first time a fully invertible neural network-based image hiding framework, FIIH, which significantly improves the performance of hiding single images, image robustness and security. Specifically, an ingenious architecture is designed to achieve invertible hiding of data by feeding only secret images into a invertible neural network. We propose hiding without adding the secret image to the cover image, which significantly improves the security of the stego image. In addition, and domain filling method is proposed for random dropout to enhance the robustness of stego image. A large number of experimental results show that this method can achieve high security, strong robustness and high invisibility of image hiding, and compared with other SOTA methods, our method achieves better results.
In the future, we will further explore the potential of invertible neural networks in multi-image hiding to achieve fully invertible multi-image hiding, i.e., hiding multiple images into another image and achieving invertible extraction. The main challenge of fully invertible multi-image hiding is how to.

\section*{Acknowledgments}
This work was supported by Open Project Program of Guangxi Key Laboratory of Digital Infrastructure (Project Number:GXDIOP2023007).

\bibliographystyle{IEEEtran}
\bibliography{ref.bib}

\begin{thebibliography}{10}
\providecommand{\url}[1]{#1}
\csname url@samestyle\endcsname
\providecommand{\newblock}{\relax}
\providecommand{\bibinfo}[2]{#2}
\providecommand{\BIBentrySTDinterwordspacing}{\spaceskip=0pt\relax}
\providecommand{\BIBentryALTinterwordstretchfactor}{4}
\providecommand{\BIBentryALTinterwordspacing}{\spaceskip=\fontdimen2\font plus
\BIBentryALTinterwordstretchfactor\fontdimen3\font minus
  \fontdimen4\font\relax}
\providecommand{\BIBforeignlanguage}[2]{{%
\expandafter\ifx\csname l@#1\endcsname\relax
\typeout{** WARNING: IEEEtran.bst: No hyphenation pattern has been}%
\typeout{** loaded for the language `#1'. Using the pattern for}%
\typeout{** the default language instead.}%
\else
\language=\csname l@#1\endcsname
\fi
#2}}
\providecommand{\BIBdecl}{\relax}
\BIBdecl

\bibitem{hsu1999hidden}
C.-T. Hsu and J.-L. Wu, ``Hidden digital watermarks in images,'' \emph{IEEE
  Transactions on image processing}, vol.~8, no.~1, pp. 58--68, 1999.

\bibitem{abraham2004significance}
A.~Abraham, M.~Paprzycki \emph{et~al.}, ``Significance of steganography on data
  security,'' in \emph{International Conference on Information Technology:
  Coding and Computing, 2004. Proceedings. ITCC 2004.}, vol.~2.\hskip 1em plus
  0.5em minus 0.4em\relax IEEE, 2004, pp. 347--351.

\bibitem{johnson1998exploring}
N.~F. Johnson and S.~Jajodia, ``Exploring steganography: Seeing the unseen,''
  \emph{Computer}, vol.~31, no.~2, pp. 26--34, 1998.

\bibitem{morkel2005overview}
T.~Morkel, J.~H. Eloff, and M.~S. Olivier, ``An overview of image
  steganography.'' in \emph{ISSA}, vol.~1, no.~2, 2005, pp. 1--11.

\bibitem{cheddad2010digital}
A.~Cheddad, J.~Condell, K.~Curran, and P.~Mc~Kevitt, ``Digital image
  steganography: Survey and analysis of current methods,'' \emph{Signal
  processing}, vol.~90, no.~3, pp. 727--752, 2010.

\bibitem{chan2004hiding}
C.-K. Chan and L.-M. Cheng, ``Hiding data in images by simple lsb
  substitution,'' \emph{Pattern recognition}, vol.~37, no.~3, pp. 469--474,
  2004.

\bibitem{tsai2009reversible}
P.~Tsai, Y.-C. Hu, and H.-L. Yeh, ``Reversible image hiding scheme using
  predictive coding and histogram shifting,'' \emph{Signal processing},
  vol.~89, no.~6, pp. 1129--1143, 2009.

\bibitem{wu2003steganographic}
D.-C. Wu and W.-H. Tsai, ``A steganographic method for images by pixel-value
  differencing,'' \emph{Pattern recognition letters}, vol.~24, no. 9-10, pp.
  1613--1626, 2003.

\bibitem{pan2011image}
F.~Pan, J.~Li, and X.~Yang, ``Image steganography method based on pvd and
  modulus function,'' in \emph{2011 International Conference on Electronics,
  Communications and Control (ICECC)}.\hskip 1em plus 0.5em minus 0.4em\relax
  IEEE, 2011, pp. 282--284.

\bibitem{das2012novel}
R.~Das and T.~Tuithung, ``A novel steganography method for image based on
  huffman encoding,'' in \emph{2012 3rd National Conference on Emerging Trends
  and Applications in Computer Science}.\hskip 1em plus 0.5em minus 0.4em\relax
  IEEE, 2012, pp. 14--18.

\bibitem{imaizumi2014multibit}
S.~Imaizumi and K.~Ozawa, ``Multibit embedding algorithm for steganography of
  palette-based images,'' in \emph{Image and Video Technology: 6th Pacific-Rim
  Symposium, PSIVT 2013, Guanajuato, Mexico, October 28-November 1, 2013.
  Proceedings 6}.\hskip 1em plus 0.5em minus 0.4em\relax Springer, 2014, pp.
  99--110.

\bibitem{fridrich2007statistically}
J.~Fridrich, T.~Pevn{\`y}, and J.~Kodovsk{\`y}, ``Statistically undetectable
  jpeg steganography: dead ends challenges, and opportunities,'' in
  \emph{Proceedings of the 9th workshop on Multimedia \& security}, 2007, pp.
  3--14.

\bibitem{hetzl2005graph}
S.~Hetzl and P.~Mutzel, ``A graph--theoretic approach to steganography,'' in
  \emph{Communications and Multimedia Security: 9th IFIP TC-6
  TC-11International Conference, CMS 2005, Salzburg, Austria, September 19--21,
  2005. Proceedings 9}.\hskip 1em plus 0.5em minus 0.4em\relax Springer, 2005,
  pp. 119--128.

\bibitem{provos2003hide}
N.~Provos and P.~Honeyman, ``Hide and seek: An introduction to steganography,''
  \emph{IEEE security \& privacy}, vol.~1, no.~3, pp. 32--44, 2003.

\bibitem{sallee2003model}
P.~Sallee, ``Model-based steganography,'' in \emph{International workshop on
  digital watermarking}.\hskip 1em plus 0.5em minus 0.4em\relax Springer, 2003,
  pp. 154--167.

\bibitem{fridrich2001detecting}
J.~Fridrich, M.~Goljan, and R.~Du, ``Detecting lsb steganography in color, and
  gray-scale images,'' \emph{IEEE multimedia}, vol.~8, no.~4, pp. 22--28, 2001.

\bibitem{hayes2017generating}
J.~Hayes and G.~Danezis, ``Generating steganographic images via adversarial
  training,'' \emph{Advances in neural information processing systems},
  vol.~30, 2017.

\bibitem{holub2014universal}
V.~Holub, J.~Fridrich, and T.~Denemark, ``Universal distortion function for
  steganography in an arbitrary domain,'' \emph{EURASIP Journal on Information
  Security}, vol. 2014, pp. 1--13, 2014.

\bibitem{holub2012designing}
V.~Holub and J.~Fridrich, ``Designing steganographic distortion using
  directional filters,'' in \emph{2012 IEEE International workshop on
  information forensics and security (WIFS)}.\hskip 1em plus 0.5em minus
  0.4em\relax IEEE, 2012, pp. 234--239.

\bibitem{pevny2010using}
T.~Pevn{\`y}, T.~Filler, and P.~Bas, ``Using high-dimensional image models to
  perform highly undetectable steganography,'' in \emph{Information Hiding:
  12th International Conference, IH 2010, Calgary, AB, Canada, June 28-30,
  2010, Revised Selected Papers 12}.\hskip 1em plus 0.5em minus 0.4em\relax
  Springer, 2010, pp. 161--177.

\bibitem{baluja2017hiding}
S.~Baluja, ``Hiding images in plain sight: Deep steganography,'' \emph{Advances
  in neural information processing systems}, vol.~30, 2017.

\bibitem{wu2018image}
P.~Wu, Y.~Yang, and X.~Li, ``Image-into-image steganography using deep
  convolutional network,'' in \emph{Advances in Multimedia Information
  Processing--PCM 2018: 19th Pacific-Rim Conference on Multimedia, Hefei,
  China, September 21-22, 2018, Proceedings, Part II 19}.\hskip 1em plus 0.5em
  minus 0.4em\relax Springer, 2018, pp. 792--802.

\bibitem{wu2018stegnet}
------, ``Stegnet: Mega image steganography capacity with deep convolutional
  network,'' \emph{Future Internet}, vol.~10, no.~6, p.~54, 2018.

\bibitem{goodfellow2014generative}
I.~Goodfellow, J.~Pouget-Abadie, M.~Mirza, B.~Xu, D.~Warde-Farley, S.~Ozair,
  A.~Courville, and Y.~Bengio, ``Generative adversarial nets,'' \emph{Advances
  in neural information processing systems}, vol.~27, 2014.

\bibitem{volkhonskiy2016generative}
D.~Volkhonskiy, B.~Borisenko, and E.~Burnaev, ``Generative adversarial networks
  for image steganography,'' 2016.

\bibitem{radford2015unsupervised}
A.~Radford, L.~Metz, and S.~Chintala, ``Unsupervised representation learning
  with deep convolutional generative adversarial networks,'' \emph{arXiv
  preprint arXiv:1511.06434}, 2015.

\bibitem{zhang2020udh}
C.~Zhang, P.~Benz, A.~Karjauv, G.~Sun, and I.~S. Kweon, ``Udh: Universal deep
  hiding for steganography, watermarking, and light field messaging,''
  \emph{Advances in Neural Information Processing Systems}, vol.~33, pp.
  10\,223--10\,234, 2020.

\bibitem{baluja2019hiding}
S.~Baluja, ``Hiding images within images,'' \emph{IEEE transactions on pattern
  analysis and machine intelligence}, vol.~42, no.~7, pp. 1685--1697, 2019.

\bibitem{jing2021hinet}
J.~Jing, X.~Deng, M.~Xu, J.~Wang, and Z.~Guan, ``Hinet: deep image hiding by
  invertible network,'' in \emph{Proceedings of the IEEE/CVF international
  conference on computer vision}, 2021, pp. 4733--4742.

\bibitem{lu2021large}
S.-P. Lu, R.~Wang, T.~Zhong, and P.~L. Rosin, ``Large-capacity image
  steganography based on invertible neural networks,'' in \emph{Proceedings of
  the IEEE/CVF conference on computer vision and pattern recognition}, 2021,
  pp. 10\,816--10\,825.

\bibitem{guan2022deepmih}
Z.~Guan, J.~Jing, X.~Deng, M.~Xu, L.~Jiang, Z.~Zhang, and Y.~Li, ``Deepmih:
  Deep invertible network for multiple image hiding,'' \emph{IEEE Transactions
  on Pattern Analysis and Machine Intelligence}, vol.~45, no.~1, pp. 372--390,
  2022.

\bibitem{shang2023robust}
F.~Shang, Y.~Lan, J.~Yang, E.~Li, and X.~Kang, ``Robust data hiding for jpeg
  images with invertible neural network,'' \emph{Neural Networks}, vol. 163,
  pp. 219--232, 2023.

\bibitem{xu2022robust}
Y.~Xu, C.~Mou, Y.~Hu, J.~Xie, and J.~Zhang, ``Robust invertible image
  steganography,'' in \emph{Proceedings of the IEEE/CVF Conference on Computer
  Vision and Pattern Recognition}, 2022, pp. 7875--7884.

\bibitem{huo2024fitting}
L.~Huo, L.~Huang, Z.~Gan, and R.~P. Chen, ``A fitting model with optimal
  multiple image hiding effect,'' \emph{Neurocomputing}, vol. 571, p. 127146,
  2024.

\bibitem{yu2020robust}
X.~Yu, K.~Chen, Y.~Wang, W.~Li, W.~Zhang, and N.~Yu, ``Robust adaptive
  steganography based on generalized dither modulation and expanded embedding
  domain,'' \emph{Signal Processing}, vol. 168, p. 107343, 2020.

\bibitem{zhang2021image}
Y.~Zhang, X.~Luo, J.~Wang, Y.~Guo, and F.~Liu, ``Image robust adaptive
  steganography adapted to lossy channels in open social networks,''
  \emph{Information Sciences}, vol. 564, pp. 306--326, 2021.

\bibitem{zhao2018improving}
Z.~Zhao, Q.~Guan, H.~Zhang, and X.~Zhao, ``Improving the robustness of adaptive
  steganographic algorithms based on transport channel matching,'' \emph{IEEE
  Transactions on Information Forensics and Security}, vol.~14, no.~7, pp.
  1843--1856, 2018.

\bibitem{zhu2018hidden}
J.~Zhu, R.~Kaplan, J.~Johnson, and L.~Fei-Fei, ``Hidden: Hiding data with deep
  networks,'' in \emph{Proceedings of the European conference on computer
  vision (ECCV)}, 2018, pp. 657--672.

\bibitem{yu2020attention}
C.~Yu, ``Attention based data hiding with generative adversarial networks,'' in
  \emph{Proceedings of the AAAI conference on artificial intelligence},
  vol.~34, no.~01, 2020, pp. 1120--1128.

\bibitem{yu2023cross}
J.~Yu, X.~Zhang, Y.~Xu, and J.~Zhang, ``Cross: Diffusion model makes
  controllable, robust and secure image steganography,'' \emph{arXiv preprint
  arXiv:2305.16936}, 2023.

\bibitem{dinh2014nice}
L.~Dinh, D.~Krueger, and Y.~Bengio, ``Nice: Non-linear independent components
  estimation,'' \emph{arXiv preprint arXiv:1410.8516}, 2014.

\bibitem{kingma2018glow}
D.~P. Kingma and P.~Dhariwal, ``Glow: Generative flow with invertible 1x1
  convolutions,'' \emph{Advances in neural information processing systems},
  vol.~31, 2018.

\bibitem{van2019reversible}
T.~F. van~der Ouderaa and D.~E. Worrall, ``Reversible gans for memory-efficient
  image-to-image translation,'' in \emph{Proceedings of the IEEE/CVF Conference
  on Computer Vision and Pattern Recognition}, 2019, pp. 4720--4728.

\bibitem{xiao2020invertible}
M.~Xiao, S.~Zheng, C.~Liu, Y.~Wang, D.~He, G.~Ke, J.~Bian, Z.~Lin, and T.-Y.
  Liu, ``Invertible image rescaling,'' in \emph{Computer Vision--ECCV 2020:
  16th European Conference, Glasgow, UK, August 23--28, 2020, Proceedings, Part
  I 16}.\hskip 1em plus 0.5em minus 0.4em\relax Springer, 2020, pp. 126--144.

\bibitem{lugmayr2020srflow}
A.~Lugmayr, M.~Danelljan, L.~Van~Gool, and R.~Timofte, ``Srflow: Learning the
  super-resolution space with normalizing flow,'' in \emph{Computer
  Vision--ECCV 2020: 16th European Conference, Glasgow, UK, August 23--28,
  2020, Proceedings, Part V 16}.\hskip 1em plus 0.5em minus 0.4em\relax
  Springer, 2020, pp. 715--732.

\bibitem{wang2020modeling}
Y.~Wang, M.~Xiao, C.~Liu, S.~Zheng, and T.-Y. Liu, ``Modeling lost information
  in lossy image compression,'' \emph{arXiv preprint arXiv:2006.11999}, 2020.

\bibitem{liu2021invertible}
Y.~Liu, Z.~Qin, S.~Anwar, P.~Ji, D.~Kim, S.~Caldwell, and T.~Gedeon,
  ``Invertible denoising network: A light solution for real noise removal,'' in
  \emph{Proceedings of the IEEE/CVF conference on computer vision and pattern
  recognition}, 2021, pp. 13\,365--13\,374.

\bibitem{zhang2019steganogan}
K.~A. Zhang, A.~Cuesta-Infante, L.~Xu, and K.~Veeramachaneni, ``Steganogan:
  High capacity image steganography with gans,'' \emph{arXiv preprint
  arXiv:1901.03892}, 2019.

\bibitem{agustsson2017ntire}
E.~Agustsson and R.~Timofte, ``Ntire 2017 challenge on single image
  super-resolution: Dataset and study,'' in \emph{Proceedings of the IEEE
  conference on computer vision and pattern recognition workshops}, 2017, pp.
  126--135.

\bibitem{russakovsky2015imagenet}
O.~Russakovsky, J.~Deng, H.~Su, J.~Krause, S.~Satheesh, S.~Ma, Z.~Huang,
  A.~Karpathy, A.~Khosla, M.~Bernstein \emph{et~al.}, ``Imagenet large scale
  visual recognition challenge,'' \emph{International journal of computer
  vision}, vol. 115, pp. 211--252, 2015.

\bibitem{kingma2014adam}
D.~P. Kingma and J.~Ba, ``Adam: A method for stochastic optimization,''
  \emph{arXiv preprint arXiv:1412.6980}, 2014.

\bibitem{boroumand2018deep}
M.~Boroumand, M.~Chen, and J.~Fridrich, ``Deep residual network for
  steganalysis of digital images,'' \emph{IEEE Transactions on Information
  Forensics and Security}, vol.~14, no.~5, pp. 1181--1193, 2018.

\end{thebibliography}

\section{Biography Section}

\vspace{11pt}

\bf{}\vspace{-33pt}
\begin{IEEEbiography}[{\includegraphics[width=1in,height=1.25in,clip,keepaspectratio]{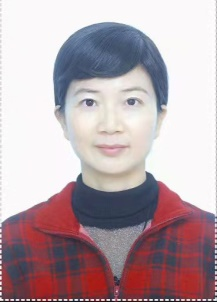}}]{Lin Huo}
received her B.Sc. degree in computer software from the Xi’an Jiaotong University, Xi’an, China. She received her M.Sc. degree in computer application and Ph.D. degree in information security from Huazhong University of Science and Technology, Wuhan, China. Since 1995, she has been a lecturer, associate professor, and professor at Guangxi University, Nanning, China. At Guangxi University, Huo’s main research areas include ciphertext retrieval, social network and distribute parallel computing. Currently, she focuses her research interests in information security, data mining, and big data.
\end{IEEEbiography}
\bf{}\vspace{-33pt}
\begin{IEEEbiography}[{\includegraphics[width=1in,height=1.25in,clip,keepaspectratio]{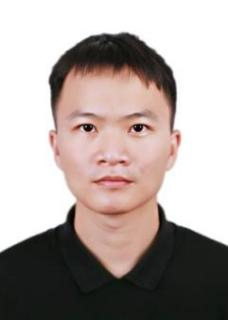}}]{Lang Huang}is currently studying for a master's degree Studied at Guangxi University, Nanning, China.His current research interests are image hiding and deep learning.
\end{IEEEbiography}
\bf{}\vspace{-33pt}
\begin{IEEEbiography}[{\includegraphics[width=1in,height=1.25in,clip,keepaspectratio]{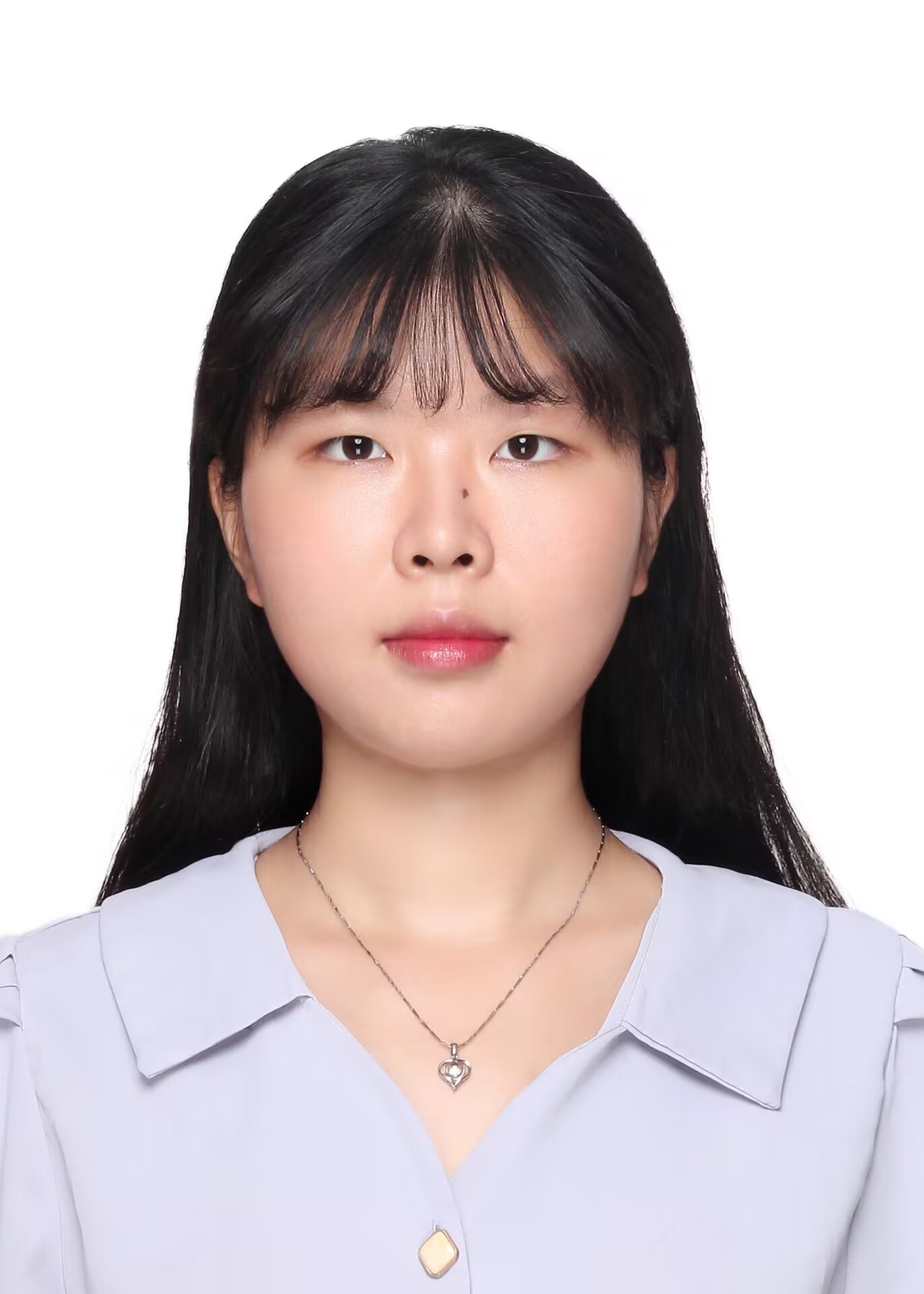}}]{Xinrong He}
a mathematics teacher at Qinzhou No. 1 Middle School, majored in mathematics and applied mathematics
\end{IEEEbiography}
\bf{}\vspace{-33pt}
\begin{IEEEbiography}[{\includegraphics[width=1in,height=1.25in,clip,keepaspectratio]{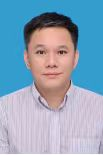}}]{Gan Zheng}received his bachelor's degree from Harbin University of Science and Technology in 2008 and his master's degree from Guangxi University in 2014. He is currently working as the deputy director of the Big Data Technology Development Department of Guangxi Zhuang Autonomous Region Information Center. His current research interests include government information system design, government information system information security.
\end{IEEEbiography}

\vfill

\end{document}